\documentclass[lettersize,journal，anonymous]{IEEEtran}
\usepackage{amsmath,amsfonts}
\usepackage[ruled,vlined,linesnumbered]{algorithm2e}

\usepackage{array}
\usepackage[caption=false,font=normalsize,labelfont=sf,textfont=sf]{subfig}
\usepackage{textcomp}
\usepackage{stfloats}
\usepackage{url}
\usepackage{verbatim}
\usepackage{graphicx}
\usepackage{cite}
\usepackage{multirow}
\usepackage{amsmath,amssymb,amsfonts}
\usepackage{graphicx}
\usepackage{xcolor}
\def\BibTeX{{\rm B\kern-.05em{\sc i\kern-.025em b}\kern-.08em
    T\kern-.1667em\lower.7ex\hbox{E}\kern-.125emX}}

\usepackage{threeparttable}
\newtheorem{definition}{Definition}
\newtheorem{theorem}{Theorem}

\usepackage{booktabs}
\usepackage{pifont}       
\usepackage{hyperref}
\begin{document}

\title{Optimization-Free Graph Embedding via Distributional Kernel for Community Detection
}

\author{
\author{\IEEEauthorblockN{ Shuaibin Song, Kai Ming Ting, Kaifeng Zhang, Tianrun Liang}\\
\IEEEauthorblockA{\textit{National Key Laboratory for Novel Software Technology} \\
\textit{Nanjing University}\\
Nanjing, China \\
\{songsb, tingkm, zhangkf, liangtr\}@lamda.nju.edu.cn}}
\thanks{This paper was produced by the IEEE Publication Technology Group. They are in Piscataway, NJ.}
\thanks{Manuscript received April 19, 2021; revised August 16, 2021.}}


\maketitle

\begin{abstract}
Neighborhood Aggregation Strategy (NAS) is a widely used approach in graph embedding, underpinning both Graph Neural Networks (GNNs) and Weisfeiler–Lehman (WL) methods. However, NAS-based methods are identified to be prone to over-smoothing—the loss of node distinguishability with increased iterations—thereby limiting their effectiveness.
This paper identifies two characteristics in a network, i.e., the distributions of nodes and node degrees that are critical for expressive representation but have been overlooked in existing methods. We show that these overlooked characteristics contribute significantly to over-smoothing of NAS-methods.
To address this, we propose a novel weighted distribution-aware kernel that embeds nodes while taking their distributional characteristics into consideration. Our method has three distinguishing features: (1) it is the first method to explicitly incorporate both distributional characteristics; (2) it requires no optimization; and (3) it effectively mitigates the adverse effects of over-smoothing, allowing WL to preserve node distinguishability and expressiveness even after many  iterations of embedding.
Experiments demonstrate that our method achieves superior community detection performance via spectral clustering, outperforming existing graph embedding methods, including deep learning methods, on standard benchmarks.
\end{abstract}


\section{Introduction}
\label{sec-introduction}

Community detection or graph clustering of an attributed network aims to partition the network into several disjoint communities in which each community has nodes satisfying structural closeness and attribute homophily \cite{chunaev2020community}. As a fundamental problem in graph data mining, it has gained traction within the research community. 

A common framework in community detection consists of two key steps. The first step is graph embedding. It aims to
 map each node and its associated neighborhood subgraph into a  vector in a new embedded space. The second  step is clustering, typically  applying an existing clustering method like spectral clustering or K-means on the set of embedded vectors produced in the first step. Evidently, the pre-requisite for successful clustering depends critically on an embedding that preserves intra-community similarity while separating inter-community differences.

However, existing embedding methods have significant challenges in achieving this aim. One major issue is over-smoothing \cite{li2018deeper}, a phenomenon characterized by the loss of node distinguishability as the number of aggregation increases, which in turn severely constrains the depth and expressiveness of of Neighborhood Aggregation Strategy (NAS)\cite{xie2020gnns}. An associated but less-studied problem is the imbalanced smoothing rates across different nodes, where high-degree or densely connected nodes tend to lose discriminability faster than sparse ones. While over-smoothing has been widely studied \cite{kipf2016semi,di2022graph,chen2020measuring}, an effective solution remains elusive. The problem of imbalanced smoothing rates has received little attention; only NDLS \cite{zhang2021node} offers a partial remedy, but it does not account for the distributional characteristics underlying this imbalance.

Furthermore, existing clustering algorithms have difficulty discovering clusters with varying densities—a well-known limitation of spectral clustering and density-based clustering (see e.g., \cite{SC-Limitations-2006,DBSCAN-IK-AAAI2019}). This limitation becomes particularly problematic when the embedded representation contains clusters of varying densities, a subtlety that existing graph embedding methods often overlook (e.g., \cite{zhang2021node,kipf2016variational,wang2019attributed,yang2020scaling}).

We argue that these challenges---of over-smoothing and embedding inadequacy of creating clusters of varying densities---stem from a fundamental oversight: existing embedding methods treat nodes by ignoring two critical distributional characteristics that substantially influence the embedding quality:

1) Distribution of node vectors that are densely populated in some regions
while others are sparse in input space. An existing NAS-method propagates and amplify this distributional characteristic
into the embedded space with increased embedding iterations.

2) Distribution of node degrees\footnote{Note that this distribution is different from the term `degree distribution' used in network science. The latter is the probability distribution of degrees over the whole network, ignoring the input space.} such that the high-degree nodes aggregate information from many neighbors, while low-degree nodes aggregate from few neighbors. This imbalance leads to different smoothing rates in different regions in the input and embedded spaces during the NAS iterations.

The failure of existing embedding methods to account for these two types of distributions in a network leads to poor community detection results.

Inspired by the success of Isolation Graph Kernel (IGK) in incorporating the distributional information in a graph to improve graph classification accuracy\cite{xu2021isolation}, we investigate whether a distribution-aware embedding approach can simultaneously address the challenges of over-smoothing, differing smoothing rates, and  clusters of varying densities in community detection. We are the first to investigate an integrated approach to deal with the impacts of these two types of distributions in community detection which  is arguably more challenging than graph classification (studied in many works on graph embedding \cite{xiao2022graph} because it is unsupervised.

The main contributions of this work are:
\begin{enumerate}
    \item Stipulating two necessary conditions of a graph embedding method (Definition \ref{Def-Graph-Embedding}) such that spectral clustering is free from the above-mentioned issue of clusters of varied densities in an embedded space.
    \item Identifying the two types of distributions in a graph that must be taken into consideration in graph embedding such that the embedded space has no clusters of varied densities. We show that existing methods are weaker because they have overlooked these distributions. We are the first to investigate these issues arising from these two types of distributions in a network.
    \item Proposing a multi-level weighted distributional kernel, called mWDK, to integrate both types of distributions as a means to graph embedding that satisfies the two conditions mentioned in the first contribution. In addition, mWDK is derived without optimization.
    \item Creating a new community detection algorithm based on mWDK  to produce better clustering outcomes, on a network with the two types of distributions, than existing graph embedding methods including deep learning based Graph Neural Networks methods.
\end{enumerate}
    \begin{figure}
    \centering
    \includegraphics[width=0.99\linewidth]{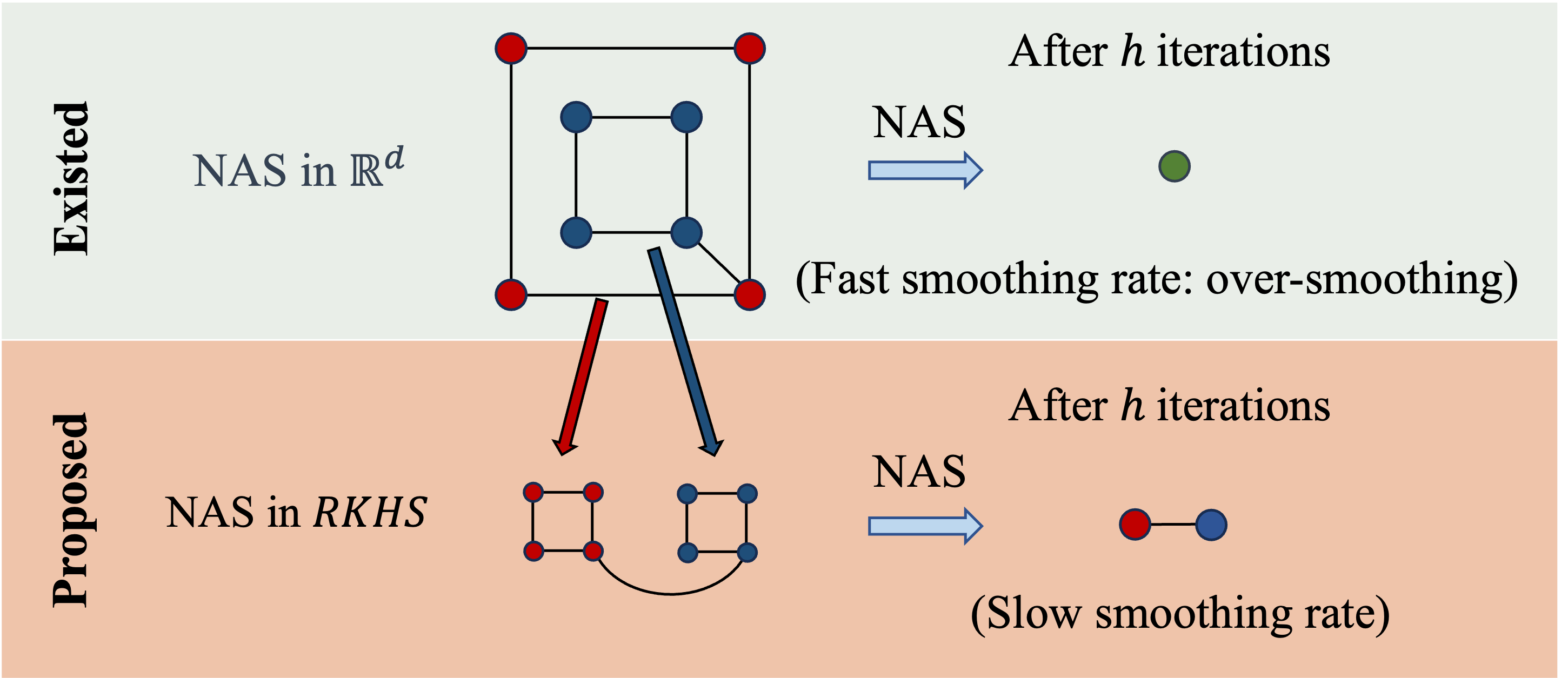}
    \caption{Neighborhood aggregation in input space (top) vs. distribution-aware space derived from the distributional kernel mWDK (bottom). Standard NAS leads to over-smoothing, merging two subgraphs into indistinguishable representations. In contrast, mWDK preserves discriminability by incorporating their distributional differences.} 
    \label{fig:idea}
\end{figure}
The proposed integrated approach is distinguished from existing approaches in the following ways:
\begin{itemize}
    \item We propose a \emph{distribution}-oriented graph embedding approach for community detection. Existing graph embedding methods are \emph{point}-oriented that do not take (a) the distribution of node vectors and (b) the distribution of node degrees into consideration.  The distribution-oriented approach via distributional kernel is a more powerful means to differentiate distinct nodes in a network with the two types of distributions, as illustrated in 
 Figure~\ref{fig:idea}. We show for the first time that WL, which is
a NAS, can maintain node discriminability during aggregations, effectively mitigating the adverse effects of over-smoothing.
    \item Isolation Kernel (IK) \cite{xu2021isolation,ting2018isolation} is used as the main tool to effectively handle  clusters of varied densities\footnote{Although IGK \cite{xu2021isolation} also employs IK, there are three key differences:
    (a) IGK uses the Isolation Forest implementation; we uses the Voronoi Diagram implementation \cite{DBSCAN-IK-AAAI2019}. (b) IGK is designed for supervised graph classification on a dataset of small graphs; but we deal with unsupervised community detection in a single large network. (c) IGK ignores the distribution of node degrees.} in the embedded space. And our method is one of the few methods that require no optimization.
    \item The key differences with all the Weisfeiler–Lehman (WL) \cite{weisfeiler1968reduction,togninalli2019wasserstein} based methods are: (i) The two types of distributions are taken into consideration. (ii) The final feature set is from the final WL iteration only, rather than a concatenation of all the WL iterations (as in IGK).
\end{itemize}

\section{Related Work}
\label{sec-related-work}
Existing unsupervised graph embedding methods can be categorized into deep learning-based, random walk-based, WL-based and those outside these three categories, as shown in Table \ref{tab:categories of embedding methods}.

\begin{table}[!ht]
\vspace{-3mm}
\caption{Unsupervised  graph embedding methods.} 
\vspace{-3mm}
     \setlength{\tabcolsep}{1.5pt} 
   \resizebox{\linewidth}{!}{
		\begin{tabular}{lrrcccr}
			\hline
			\multirow{2}{*}{Category}         & \multirow{2}{*}{Methods}         & Optimization                          & \multirow{2}{*}{NAS} & \multicolumn{2}{c}{Distribution of} & \multirow{2}{*}{Time complexity}\\ \cline{5-6} 
			& & Free & &Nodes & Degrees\\ \hline
			   		
			\multirow{4}{*}{Deep learning}         & GAE \cite{kipf2016variational}                                                                      &   \ding{55} &   \ding{51}&   \ding{55} &   \ding{55}&  Not provided  \\
			& MGAE \cite{wang2017mgae}                                                                        &   \ding{55} &   \ding{51} &   \ding{55} &   \ding{55} &  Not provided  \\
			& DAEGC \cite{wang2019attributed}                                               &   \ding{55} &   \ding{51}&   \ding{55} &   \ding{55} & Not provided \\

           & MAGI \cite{liu2024revisiting} &   \ding{55} &   \ding{51}&   \ding{55} &   \ding{55} & Not provided \\
			\hline
			\multirow{5}{*}{WL}      			& WWL \cite{togninalli2019wasserstein}                                                           &   \ding{55} &   \ding{51} &   \ding{55}  &   \ding{55}&$O(n^{3}\log{(n)}+mdh)$ \\
   & WL-f \cite{weisfeiler1968reduction,togninalli2019wasserstein}      &   \ding{51}  &   \ding{51} &   \ding{55} &   \ding{55} &$O(mdh)$\\
   & SGC \cite{wu2019simplifying}, AGC \cite{zhang2023adaptive} &   \ding{51} &   \ding{51} &   \ding{55} &   \ding{55} &  $O(mdh)$ \\  
			& IGK \cite{xu2021isolation}                                                                       &   \ding{51} &   \ding{51} &   \ding{51} &   \ding{55}  &$O(\psi t(n+mh))$  \\
    &\textbf{mWDK(ours)}               &   \ding{51} &   \ding{51} &   \ding{51}&   \ding{51}  &$O(\psi t h(n+m))$\\
     \hline
			\multirow{2}{*}{Random walk} & TADW \cite{yang2015network}                                         &   \ding{55} &   \ding{55} &   \ding{55}  &   \ding{55} & $O(mk + nftk + nk^{2} ) $\\
			& PANE \cite{yang2020scaling}                                   &   \ding{55} &   \ding{55} &   \ding{55} &   \ding{55}&$O(d(m+nb)) \cdot \log{\frac{1}{\varepsilon } }$  \\
			\hline
     \multirow{2}{*}{Others} 
			& NDLS-F \cite{zhang2021node}                                                     &   \ding{51} &   N/A &   \ding{55} &   \ding{51} &$O(m(d+c)h)$ \\
			&GSNN \cite{wijesinghe2022new}                     &   \ding{51} &    N/A &   \ding{55} &   \ding{55} &$O(kmdf)$\\
			\hline
		\end{tabular}}
	\begin{tablenotes}
		\footnotesize
		\item $n$, $m$ and $d$ denote the numbers of nodes, edges and dimensions, respectively.  $h$ is the iteration parameter of WL. $\psi$ and $t$ are the IK parameters. $k$ is the number of GNN layers. $b$ and $f$ are the number of dimensions of input and output features, respectively. $c$ is the number of class.
	\end{tablenotes}

	\label{tab:categories of embedding methods}
 \vspace{-2mm}
\end{table}


Neighborhood aggregation strategy (NAS) \cite{xie2020gnns} is a common approach for many methods.
NAS refers to ways to aggregate information from a node's neighborhood in order to produce an embedding for the node. The aggregation function may differ from one method to the other \cite{Aggregation-NIPS2020}.

For a given attributed network, 
let $\mathbf{X}$ and $\mathbf{A}$ denote the feature matrix of node vectors and the adjacency matrix of the network, respectively. The NAS process at the $h$-th iteration can be formally expressed as:
\begin{equation}
\mathbf{X}^h = \mathbf{W} \mathbf{X}^{h-1},
\end{equation}
\noindent
where $\mathbf{W}$ denotes the aggregation weights derived from the normalized adjacency matrix $\mathbf{A}$. 


Intuitively, as $h$ increases, NAS enables the connected nodes to become increasingly similar in the embedded space,  producing  high-quality node embeddings in a homophilic network \cite{chunaev2020community}. However, it falls short in addressing a network containing numerous heterophilic nodes, where a node and its neighboring nodes may belong to different communities. This leads to a phenomenon known as over-smoothing, which will be further discussed in Section \ref{sec-over-smoothing}.

In the next three subsections, we will present two types of NAS methods, i.e., deep learning based methods and Weisfeiler–Lehman methods, and a non-NAS method based on random walk.

\subsection{Deep learning based methods}
Deep learning methods have been extensively researched for community detection, showing competitive results, such as DMoN \cite{tsitsulin2023graph}, GAE \cite{tu2021siamese}, VGAE \cite{kipf2016variational}, AGE \cite{cui2020adaptive} and DAEGC \cite{wang2019attributed}.
The typical model is Graph Convolutional Network (GCN) \cite{kipf2016semi}, utilizing a Laplacian operator for neighbor aggregation. DMoN is specially designed for unsupervised embedding. GAE, VGAE and AGE are based on autoencoders. DAEGC \cite{wang2019attributed} is reported to achieve SOTA performance in clustering by incorporating attention mechanism on GAE. MAGI \cite{liu2024revisiting} uses modularity maximization as a pretext task for contrastive representation learning. It employs a two-stage random walk with mini-batch training to enable scalable and effective graph clustering.

However, these methods are computationally expensive because they require training on a large number of parameters.

\subsection{Weisfeiler–Lehman methods}

The Weisfeiler–Lehman (WL) algorithm \cite{weisfeiler1968reduction} was initially  proposed to detect isomorphic labelled graphs. It has been modified as a graph kernel method via a subtree neighborhood \cite{shervashidze2011weisfeiler} and further extended into continuous versions in order to apply to attributed graphs \cite{morris2016faster,togninalli2019wasserstein}. A direct treatment is to iteratively update a current node's representation by averaging over the node's neighborhood.

Let $\mathcal{G}=(\mathcal{V},\mathcal{E})$ be an attributed network, where $\mathcal{V}$ and $\mathcal{E}$ denote the sets of nodes and edges, respectively; and $\mathcal{N}(v)$ be the set of node vectors in the 1-hop neighborhood of node vector $v \in \mathcal{V}$, excluding $v$. Further let $v^i$ be the node vector of $v$ at iteration $i$ of WL; and  $v^0=v$. The terms `node vectors' and `points' in $\mathbb{R}^d$ are used interchangeably hereafter.

\begin{definition}

A Weisfeiler–Lehman (WL) scheme  that embeds node vector $v$ associated with neighborhood subgraph $\mathcal{G}$, denoted as $\mathcal{G}(v)$,  is defined \cite{togninalli2019wasserstein} as:
	\begin{equation}
		\begin{aligned}
			\mathbf{v}^{(h)}=[v^{0},v^{1},\cdots,v^{h}], 
		\end{aligned}
	\end{equation}
\noindent
where $v^{h}$ represents the embedding of $\mathcal{G}(v)$ at iteration $h$:
\begin{equation}
	v^{h}=\frac{1}{2}\left(v^{(h-1)}+\frac{1}{|\mathcal{N}(v)|} \sum_{u \in \mathcal{N}(v)}  u^{(h-1)}\right).\label{eqn-WL} 
\end{equation}

\end{definition}

From this perspective, both WL and GCN are implementations of NAS. The differences are: (i)  normalizations; and (ii) GCN learns the weights but WL uses a simple weighted averaging without optimization. Optimization-free versions of GCN, exemplified by  AGC \cite{zhang2023adaptive} and SGC \cite{wu2019simplifying}, employ a  customized convolution kernel instead of learning. We examine the impact of these differences in an ablation study.

WWL \cite{togninalli2019wasserstein} integrates the WL scheme with Wasserstein distance (WD), incurring high time costs due to WD and lacking an explicit feature map. IGK \cite{xu2021isolation} reduces the time cost with an explicit feature map from Isolation Kernel \cite{ting2018isolation}, and it takes the distribution of node vectors into consideration, whereas WWL does not. 

\subsection{Random Walk based methods}

Motivated by the success of random walk, particularly the effective application of word2vec \cite{mikolov2013efficient} in natural language processing, approaches based on random walk are widely used for graph embedding. For instance, DeepWalk \cite{perozzi2014deepwalk} and Node2vec \cite{grover2016node2vec} convert nodes in a network into node sequences by sampling from random walks, and then learn an embedding using skip-grams \cite{mikolov2013efficient}. However, these models 
could not deal with attributed graphs. 
TADW \cite{yang2015network} attempts to address this limitation by injecting node features into the matrix factorization process in DeepWalk. PANE \cite{yang2020scaling} achieves this by treating  node features as visual nodes and adding them to the original graph to reconstruct a new graph. 
These methods capture the similarity between nodes relying on their random walk sampling sequences, which differ from the NAS approach.

\section{Over-smoothing: a key issue of NAS}
\label{sec-over-smoothing}

NAS is perceived to have a key issue dubbed `over-smoothing' \cite{li2018deeper}. It imposes a serious negative effect because any two NAS-embedded nodes in a graph become increasingly similar as the number of iterations increases in NAS. The over-smoothing effect has been rigorously defined \cite{rusch2023survey}. Here we use a simpler definition, so that the relative robustness of two embedding methods can be compared with reference to a same similarity threshold,    as follows.

\begin{definition}
Two  points $v$ and $u$ in graph $\mathcal{G}$, with their respective embedded points $v^h$ and $u^h$ derived from an embedding method of $h$ iterations,  are said to be $\mu$-similar if their similarity $\kappa(v^h,u^h) \ge \mu$, where $\mu \in (0, 1]$.
\label{def-mu-similar}
\end{definition}

\begin{definition}
An embedding method of $h$ iterations has an over-smoothing effect at $\mu$ level if all pairs of points $v$ and $u$ in graph $\mathcal{G}$ are $\mu$-similar. 
\label{def-oversmoothing}
\end{definition}

Given two embedding methods, a method is more robust to the over-smoothing effect  than the other if the former has a higher $h$ than the latter to reach the same $\mu$ level, i.e., the former has a slower smoothing rate than the latter.

A growing body of research has shown the impact of over-smoothing on graph embedding, either theoretically or experimentally \cite{kipf2016semi,di2022graph,chen2020measuring}. 
Several studies  have employed  graph data augmentation such as DropNode \cite{huang2018adaptive}, DropEdge \cite{rong2019dropedge}, Dropout \cite{srivastava2014dropout}, GRAND \cite{feng2020graph} and EP \cite{liu2022alleviating} to decrease the rate of over-smoothing. Some works alleviate the over-smoothing by enhancing the aggregation mechanism. These include NDLS \cite{zhang2021node}, GCNII \cite{chen2020simple} and JK-Net \cite{xu2018representation}.

All the above  methods do not exploit the distribution information of attributed graphs---a key in addressing the over-smoothing effect we found here. This omission could yield clusters of hugely different densities in the embedded space, adversely affecting clustering algorithms like spectral clustering,   as  discussed in Section \ref{sec-introduction}.

In the next three sections, we will present the idea of using distributional information in graph embedding to address the over-smoothing. Section \ref{sec-community-detection} describes the general idea,  Section~\ref{sec-WDK} provides the proposed weighted  distributional kernel method, and Section \ref{sec-mWDK-impact} presents a demonstration of the effect and a theoretical analysis.
\section{Distribution-oriented approach}
\label{sec-community-detection}
Community detection is a clustering problem in a network with the aim to identify communities as clusters, having edges denoting the connectivity of nodes within a cluster/community, while some edges obscure the boundary between any two clusters/communities. It is formally defined as follows:




\begin{definition}
	A community detection task in an unweighted network $\mathcal{G}= (\mathcal{V},\mathcal{E})$ aims to cut the edges which obscure the boundary between any two different communities,  where $\mathcal{V}$ is set of points/nodes in $\mathbb{R}^d$, and $\mathcal{E}$ is a set of edges which connect pairs of points in $\mathcal{V}$. 
	\label{Def-Community-Detection}
\end{definition}

Given Definition \ref{Def-Community-Detection}, the idea is to make full use of the connectivity already given in the network $\mathcal{G}$ for each community in order to find and cut the unwanted edges that obscure the boundary between different communities. Thus the top-level view of the approach is to \emph{first accentuate each community, and then sever the unwanted edges between different communities}.

\subsection{Distribution-oriented graph embedding}

To accentuate the connectivity within a community, we propose to use a distribution-oriented approach to graph embedding. It treats all points connected in a subgraph as a set of independent and identically distributed (i.i.d.) sampled points from an unknown distribution. The subgraph considered is formally defined as follows:

\begin{definition}
	An $h$-subgraph $\mathcal{G}^h(v) = (\mathbf{V}_v,\mathbf{E})$ is a subgraph rooted at point $v$, where the shortest path between any point $u \in \mathbf{V}_v$ and $v$ has length $\le h$, and $h \in \mathbb{N}$ is the maximum depth of the subgraph. 
 The average node degree in $\mathcal{G}^h(v)$ is defined as $\frac{1}{|\mathbf{V}_v|}\sum_{u \in \mathbf{V}_v} deg(u)$, where $deg(u)$ is the degree of $u$.
\end{definition}

The points in $\mathbf{V}_v$ is then treated as a set of i.i.d. sampled points from an unknown distribution $\mathcal{P}_{\mathbf{V}_v}$. This assumption allows us to use a distributional kernel to map $\mathcal{P}_{\mathbf{V}_v}$ to a point in RKHS (reproducing kernel Hilbert space). 

A community of $m$ points has $m$ $h$-subgraphs, where $\mathbf{V}_i, i=1,\dots,m$, are sample sets generated from the same distribution $\mathcal{P}_\mathbf{V}$. After mapping with a distributional kernel, the $m$ sample sets become $m$ points in a cluster in RKHS. Different communities in a network are mapped to different clusters in RKHS.
An existing clustering algorithm for a set of points can then be used to discover the clusters.

\emph{The key difference between the proposed approach and existing approaches is that the former takes the distribution into consideration; whereas the latter does not}. 

\subsection{The aim of the proposed approach}
A good method of graph embedding for community detection shall achieve the following aim: 
\begin{definition}
	Graph embedding for a community detection task in a network $\mathcal{G}= (\mathcal{V},\mathcal{E})$ aims to (i) accentuate the similarity $S(\mathcal{C})$ within each community $\mathcal{C}$ 
	and (ii) maintain the similarity $S(\mathcal{C}_i, \mathcal{C}_j)$   at a low level between two different communities $\mathcal{C}_i \ne \mathcal{C}_j$. 
	\label{Def-Graph-Embedding}
\end{definition}

The similarities $S(\cdot,\cdot)$ and $S(\cdot)$ computed using a distributional kernel are given in Section \ref{sec-similarities}. 

The first condition is to ensure that every community has approximately the same similarity (equivalent to approximately the same density in the embedded space). Many clustering algorithms  have difficulty identifying clusters with varied densities. It is a known fundamental limitation of spectral clustering \cite{SC-Limitations-2006} and density-based clustering \cite{Density-issue-PRJ2016}. The condition allows these clustering algorithms to be used for community detection without such difficulty. 

The second condition  enables the obscure boundaries between communities, stated in Definition \ref{Def-Community-Detection}, to be found more easily in the embedded space by a clustering algorithm.

The proposed method is tasked to satisfy both conditions, stated in Definition \ref{Def-Graph-Embedding}. Its details are in the next section.

\section{Weighted Distributional kernel for $h$-subgraph embedding}
\label{sec-WDK}
\begin{figure*}[h]
    \centering
    \includegraphics[width=\textwidth]{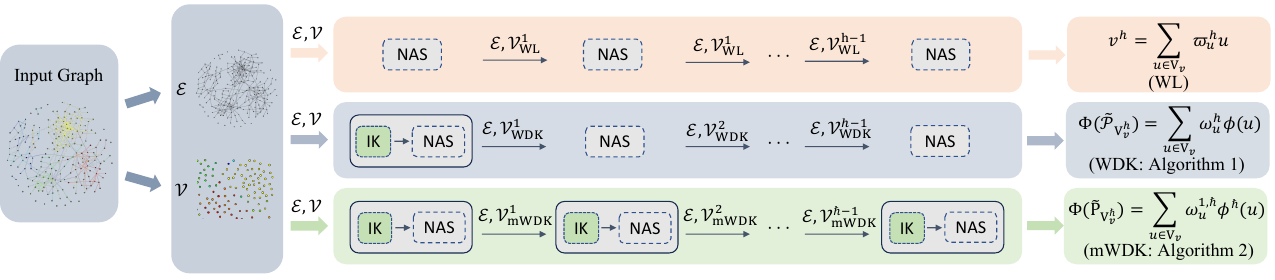}
    \caption{Illustrations of WL, WDK and mWDK embedding methods: $v_{\mathrm{WL}}^{h}=\bigcup_{u \in \mathcal{V}_v} v^{h}$, $\ \mathcal{V}_{\mathrm{WDK}}^{h}=\bigcup_{u \in \mathcal{V}_v} \Phi\left(\tilde{\mathcal{P}}_{\mathbf{V}_{u}^{h}}\right)$ and $\mathcal{V}_{\mathrm{mWDK}}^{\hslash}=\bigcup_{u \in \mathcal{V}_v} \Phi\left(\tilde{\mathsf{P}}_{\mathbf{V}_{u}^{\hslash}}\right)$.}
    \label{fig:Embedding-illustration}
\end{figure*}
Given an attributed network $\mathcal{G}=(\mathcal{V},\mathcal{E})$, the embedding of $h$-subgraph $\mathcal{G}^h(v)=(\mathbf{V}_v,\mathbf{E})$ for all $v \in \mathcal{V}$, 
based on a Weisfeiler–Lehman (WL) scheme \cite{togninalli2019wasserstein} defined in Eq~\ref{eqn-WL},
can be re-expressed as:
\begin{equation}
	v^{h}= \sum_{u \in \mathbf{V}_v} \varpi^{h}_u \times u, \label{eqn-weighted-distribution1} 
\end{equation}
\noindent
where $ \varpi^{h}_u$ is the weight of $u$ in $\mathcal{G}^h(v)$, and $\sum_{u\in \mathbf{V}_v}  \varpi^{h}_u = 1$. 

When the above WL-embedding is viewed as giving rise to a weighted distribution of points in $\mathbf{V}_v$, i.e., $\widetilde{\mathcal{P}}_{\mathbf{V}_v}$,
the kernel mean embedding\footnote{The ordinary kernel mean embedding for the (unweighted) distribution $\mathcal{P}_{\mathbf{V}_v}$ is defined as \cite{KernelMeanEmbedding2017}: $\Phi(\mathcal{P}_{\mathbf{V}_v})= \frac{1}{|\mathbf{V}_v|}\sum_{u \in \mathbf{V}_v} \phi(u)$.}  for $\widetilde{\mathcal{P}}_{\mathbf{V}_v}$ based on a base kernel $\kappa(u,v) = \left< \phi(u), \phi(v) \right>$ can be then expressed as:
\begin{equation}
	\Phi(\widetilde{\mathcal{P}}_{\mathbf{V}_v})= \sum_{u \in \mathbf{V}_v}  \omega^{h}_u \times \phi(u).\label{eqn-mean-map} 
\end{equation}
\noindent
where $\omega^{h}_u$ is the weight of $\phi(u)$ in the feature space of $\kappa$, and $\sum_{u\in \mathbf{V}_v} \omega^{h}_u = 1$.

This is achieved by mapping all $v \in \mathcal{V}$ to $\phi(v)$ in RKHS induced by the kernel $\kappa$, before applying WL using Eq \ref{eqn-WL} in RKHS for each $h$-subgraph $\mathcal{G}^h(v)=(\mathbf{V}_v,\mathbf{E})$ for all $v \in \mathcal{V}$. 

As a result, a network of $n=|\mathcal{V}|$ node vectors gives rise to a set of $n$ embedded vectors $\Phi(\widetilde{\mathcal{P}}_{\mathbf{V}_v})$.
Its corresponding distributional kernel for two distributions $\widetilde{\mathcal{P}}_{\mathbf{V}_v}$ and $\widetilde{\mathcal{P}}_{\mathbf{W}}$ (of two $h$-subgraphs $\mathcal{G}^h(v) = (\mathbf{V}_v,\mathbf{E})$ and $\mathcal{G}^h(w)= (\mathbf{W},\mathbf{F})$) is:
\begin{equation}
    \mathcal{K}(\widetilde{\mathcal{P}}_{\mathbf{V}_v}, \widetilde{\mathcal{P}}_{\mathbf{W}}) = \left<\Phi(\widetilde{\mathcal{P}}_{\mathbf{V}_v}),\Phi(\widetilde{\mathcal{P}}_{\mathbf{W}})\right>. 
\end{equation}

We call the above WL-induced kernel: Weighted Distributional Kernel (WDK).

Instead of using the WDK embedding as shown in Eq (\ref{eqn-mean-map}), one may use $h=1$ (i.e., $1$-subgraph) to produce the base WDK; and then re-apply the base WDK on top of the previous base WDK multiple times to construct multi-level WDK. The details are given as follows.

Let $\widetilde{\mathsf{P}}_{\mathbf{V}_v^\hslash}$ be the weighted distribution of $\mathbf{V}_v^\hslash$, where $\mathbf{V}_v^\hslash$ is a set of points  and $\mathbf{E}^\hslash$ is set of edges in $(1,\hslash)$-subgraph $(\mathbf{V}_v^\hslash,\mathbf{E}^\hslash)$ in level-$\hslash$ embedded space. The input space is level-0 embedded space.


The kernel mean embedding of level-$\hslash$ WDK using $\hslash$ levels of $(1,\hslash)$-subgraphs and $\phi^\hslash$ feature maps of kernel $\kappa$
is defined as:
\begin{equation}
\Phi(\widetilde{\mathsf{P}}_{\mathbf{V}_v^\hslash})= \sum_{u  \in \mathbf{V}_v^\hslash} \omega^{1,\hslash}_{u} \times \phi^\hslash(u).
 \label{eqn-weighted-distribution} 
\end{equation}

In other words,  mWDK applies $\hslash$ number of feature maps of $\kappa$ at $\hslash$ levels of $(1,\hslash)$-subgraphs; whereas the WDK (defined in Eq (\ref{eqn-mean-map})) applies one feature map of $\kappa$ to one $h$-subgraph.
Using mWDK with $\hslash$, the corresponding distributional kernel for two $(1,\hslash)$-subgraphs $\mathcal{G}^{1,\hslash}(v) = (\mathbf{V}_v^\hslash,\mathbf{E}^\hslash)$ and $\mathcal{G}^{1,\hslash}(w)= (\mathbf{W}^\hslash,\mathbf{F}^\hslash)$ is:
\begin{equation}
    \mathcal{K}(\widetilde{\mathsf{P}}_{\mathbf{V}_v^\hslash}, \widetilde{\mathsf{P}}_{\mathbf{W}^\hslash}) = \left<\Phi(\widetilde{\mathsf{P}}_{\mathbf{V}_v^\hslash}),\Phi(\widetilde{\mathsf{P}}_{\mathbf{W}^\hslash})\right>. 
\end{equation}

Figure \ref{fig:Embedding-illustration} provides an illustrative comparison of the above three embedding methods, i.e., WL, WDK, mWDK. Their algorithms are provided in the next subsection. 

\subsection{Algorithms for WDK, mWDK and WL} The community detection algorithms using WDK and mWDK are given in Algorithms \ref{alg:WDK} and \ref{alg:mWDK}, respectively. They are direct implementations of Eq (\ref{eqn-mean-map}) and Eq (\ref{eqn-weighted-distribution}) to produce the embedded spaces before applying spectral clustering (SC).

\begin{algorithm}[ht]
	\SetAlgoLined
	\SetKwInOut{Input}{Input}\SetKwInOut{Output}{Output}
	\Input{$\mathcal{G}=(\mathcal{V},\mathcal{E})$ - given attributed network \& $\mathcal{V} \subset \mathbb{R}^d$, $k$ - number of clusters, $h$ - level of subgraph}
	\Output{$\mathcal{G}_{\mathcal{C}_j}, j=1,...,k$}
	\BlankLine
		
	Produce $\phi(\cdot\mid \mathcal{V})$ from $\mathcal{V}$, where $\phi(\cdot)$ is the feature map of a base kernel $\kappa$\;
		
	
$\forall v \in \mathcal{V}$, map each $h$-subgraph $\mathcal{G}^h(v)=(\mathbf{V}_v,\mathbf{E})$   
	via WDK:
	$\Phi(\widetilde{\mathcal{P}}_{\mathbf{V}_v})= \sum_{u \in {\mathbf{V}_v}} \omega^{h}_u \times \phi(u)$ [Eq (\ref{eqn-mean-map})]\;
	$\mathcal{D}=\bigcup_{v\in \mathcal{V}}\Phi(\widetilde{\mathcal{P}}_{\mathbf{V}_v})$;
		
	Produce $k$ clusters $\mathcal{C}_j, j=1,\dots,k$ from $\mathcal{D}$ using SC\;
	Community $\mathcal{G}_{\mathcal{C}}$ is a subgraph of $\mathcal{G}$ which contains a subset of points $v \in \mathcal{V}$ correspond to $\Phi(\widetilde{\mathcal{P}}_{\mathbf{V}_v})$  in $\mathcal{C}$\;
	\Return Communities $\mathcal{G}_{\mathcal{C}_j}, j=1,\dots,k$\;
	\caption{Community detection with WDK}
	\label{alg:WDK}
\end{algorithm}

\begin{algorithm}[ht]
	\SetAlgoLined
	\SetKwInOut{Input}{Input}\SetKwInOut{Output}{Output}
	\Input{$\mathcal{G}=(\mathcal{V},\mathcal{E})$ - given attributed network \& $\mathcal{V} \subset \mathbb{R}^d$, $k$ - number of clusters,  $\hslash $ - level of mWDK}
	\Output{$\mathcal{G}_{\mathcal{C}_j}, j=1,...,k$}
	\BlankLine
	Initialize $\mathcal{D}^{1}=\mathcal{V}$;
		
	\For{each $i \in [0,\hslash]$}{
		Produce $\phi^i(\cdot\mid \mathcal{D}^{i})$ from $\mathcal{D}^{i}$, where $\phi^i(\cdot)$ is the feature map of a base kernel $\kappa$ at level-$i$\;
		$\forall v \in \mathcal{D}^{i}$, map each $(1,i)$-subgraph $\mathcal{G}^{1,i}(v)=(\mathbf{V}_v^i,\mathbf{E}^i)$   
		via mWDK at  level-$i$:\hspace{4cm}.
		$\Phi(\widetilde{\mathsf{P}}_{\mathbf{V}_v^i})= \sum_{u \in \mathbf{V}_v^i} \omega^{1,i}_{u} \times \phi^i(u)$ [Eq (\ref{eqn-weighted-distribution})]\;
		$\mathcal{D}^{i+1}=\bigcup_{v \in \mathcal{D}^{i}}\Phi(\widetilde{\mathsf{P}}_{\mathbf{V}_v^i})$;
	}	
	Produce $k$ clusters $\mathcal{C}_j, j=1,\dots,k$ from $\mathcal{D}^{\hslash+1}$ using SC\;
	Community $\mathcal{G}_{\mathcal{C}}$ is a subgraph of $\mathcal{G}$ which contains a subset of points $v \in \mathcal{V}$ correspond to $\Phi(\widetilde{\mathsf{P}}_{\mathbf{V}_v^{\hslash}})$  in $\mathcal{C}$\;
	\Return Communities $\mathcal{G}_{\mathcal{C}_j}, j=1,\dots,k$\;
	\caption{Community detection with mWDK}
	\label{alg:mWDK}
\end{algorithm}

In each algorithm, the clustering outcome is the communities of nodes ($v \in \mathcal{V}$) in the input space which correspond to the discovered clusters of embedded vectors  in the embedded space, where $\Phi(\widetilde{\mathcal{P}}_{\mathbf{V}_v})$ and $\Phi(\widetilde{\mathsf{P}}_{\mathbf{V}_v^\hslash})$ are the embedded vectors for $\mathcal{G}^h(v)$ and $\mathcal{G}^{1,\hslash}(v)$ in WDK and mWDK, respectively. For brevity, we denote $h$ as $\hslash$ when referring to mWDK hereafter, unless a distinction is required.

The community detection algorithm using WL (rather than mWDK or WDK) is the same as Algorithm \ref{alg:WDK} by replacing lines 1-3 with $\mathcal{D}=\bigcup_{v\in \mathcal{V}} v^{h}$, and line~5 becomes: `Community $\mathcal{G}_{\mathcal{C}}$ is a subgraph of $\mathcal{G}$ which contains a subset of points $v \in \mathcal{V}$ correspond to $v^{h}$ [defined in Eq (\ref{eqn-weighted-distribution1})] in cluster $\mathcal{C}$'.


\subsection{Computing similarities of communities}
\label{sec-similarities}

Let  $\mathcal{C} = (\mathcal{V}_\mathcal{C}, \mathcal{E}_\mathcal{C})$ be a community in a network $\mathcal{G}$. The similarity between two communities $\mathcal{C}_1$ and $\mathcal{C}_2$ computed using WDK $\mathcal{K}$ with $h$ iterations is given as:
\begin{equation}
    S_h(\mathcal{C}_1,\mathcal{C}_2) = \frac{1}{|\mathcal{C}_1||\mathcal{C}_2|} \sum_{v \in \mathcal{V}_{\mathcal{C}_1}, w \in \mathcal{V}_{\mathcal{C}_2}} \mathcal{K}(\widetilde{\mathcal{P}}_{\mathbf{V}_v}, \widetilde{\mathcal{P}}_{\mathbf{W}}).
\end{equation}
The similarity within a community $S(\mathcal{C})$ can then be computed by setting $\mathcal{C}_1 = \mathcal{C}_2$ in the above computation. 

In the same manner, the similarities of communities based on mWDK can be computed by simply replacing $\mathcal{K}(\widetilde{\mathcal{P}}_{\mathbf{V}_v}, \widetilde{\mathcal{P}}_{\mathbf{W}})$ with $\mathcal{K}(\widetilde{\mathsf{P}}_{\mathbf{V}_v^\hslash}, \widetilde{\mathsf{P}}_{\mathbf{W}^\hslash})$ in the above formulation. And the similarities of communities corresponding to WL  is calculated by using a linear kernel for $v^{h}$ and $w^{h}$ (in place of $\mathcal{K}(\widetilde{\mathcal{P}}_{\mathbf{V}_v}, \widetilde{\mathcal{P}}_{\mathbf{W}})$).

\begin{figure}[t]
	\centering
\includegraphics[width=0.9\linewidth,height=0.8\linewidth]{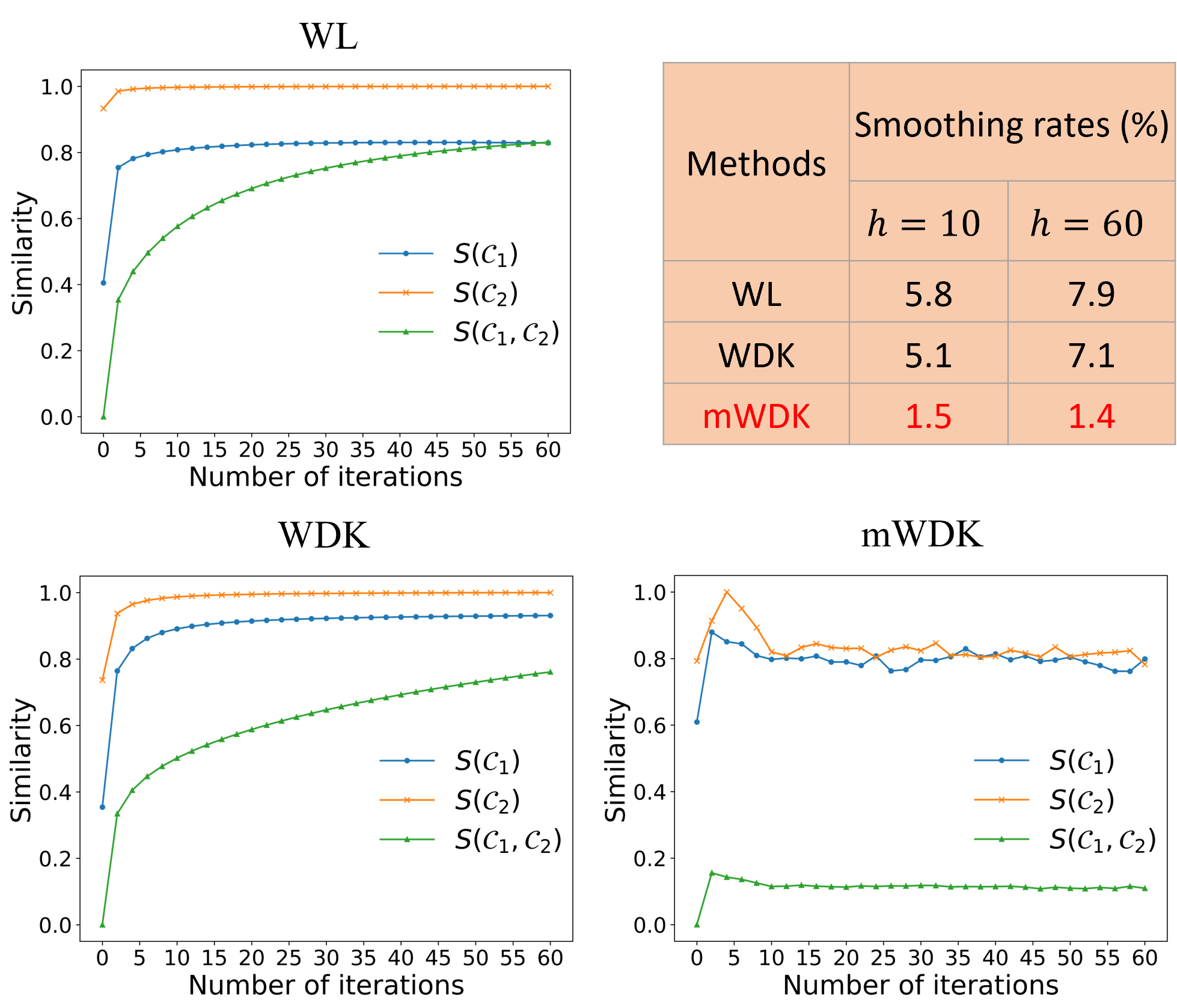}
	\caption{Similarity of WL, WDK, mWDK on the UE dataset having two communities $\mathcal{C}_1$ \& $\mathcal{C}_2$ (described in Section \ref{Sec-Datasets}) with a uniform distribution of node degrees but an imbalanced  distribution of nodes. The top-right table shows a comparison of smoothing rates of WL, WDK \& mWDK. The smoothing rate at $h$-th iteration for $\mathcal{C}_1$ and $\mathcal{C}_2$  is computed as:
    $R_h(\mathcal{C}_1, \mathcal{C}_2) =  \frac{S_h(\mathcal{C}_1,\mathcal{C}_2) - S_0(\mathcal{C}_1,\mathcal{C}_2)}{h}$,
where $S_0(\mathcal{C}_1,\mathcal{C}_2)$ is the similarity in the input space. 
}\label{fig:sim}
\end{figure}

\section{{mWDK} produces well-separated clusters of equal density in embedded space}
\label{sec-mWDK-impact}


Here we first use two example datasets to demonstrate that mWDK produces well-separated clusters of equal density in the embedded space. Then, we explain the reasons why mWDK has this ability, whereas WDK does not.

\vspace{2mm}
\noindent
\textbf{Dataset having two clusters with unequal node densities but equal node degree}.
Figure \ref{fig:sim}  shows the trends of  $S(\mathcal{C}_i)$ and  $S(\mathcal{C}_1,\mathcal{C}_2)$ curves as $h$ increases for WL, WDK \& mWDK on this dataset. The key difference is shown in the $S(\mathcal{C}_1,\mathcal{C}_2)$ curves which indicate that the smoothing rate is high if $S(\mathcal{C}_1,\mathcal{C}_2)$ gets larger as $h$ increases. 
WL has the typical  over-smoothing effect, i.e., the two clusters become increasingly similar as $h$ increases, shown in the WL subfigure. 
WDK reduces the over-smoothing effect of WL  at a slower rate, as shown in the WDK subfigure. 
In sharp contrast, mWDK has almost no smoothing effect,
as shown in the mWDK subfigure with a flat $S(\mathcal{C}_1,\mathcal{C}_2)$ curve. 



\begin{figure}[t]
	\centering
	\includegraphics[width=\linewidth]{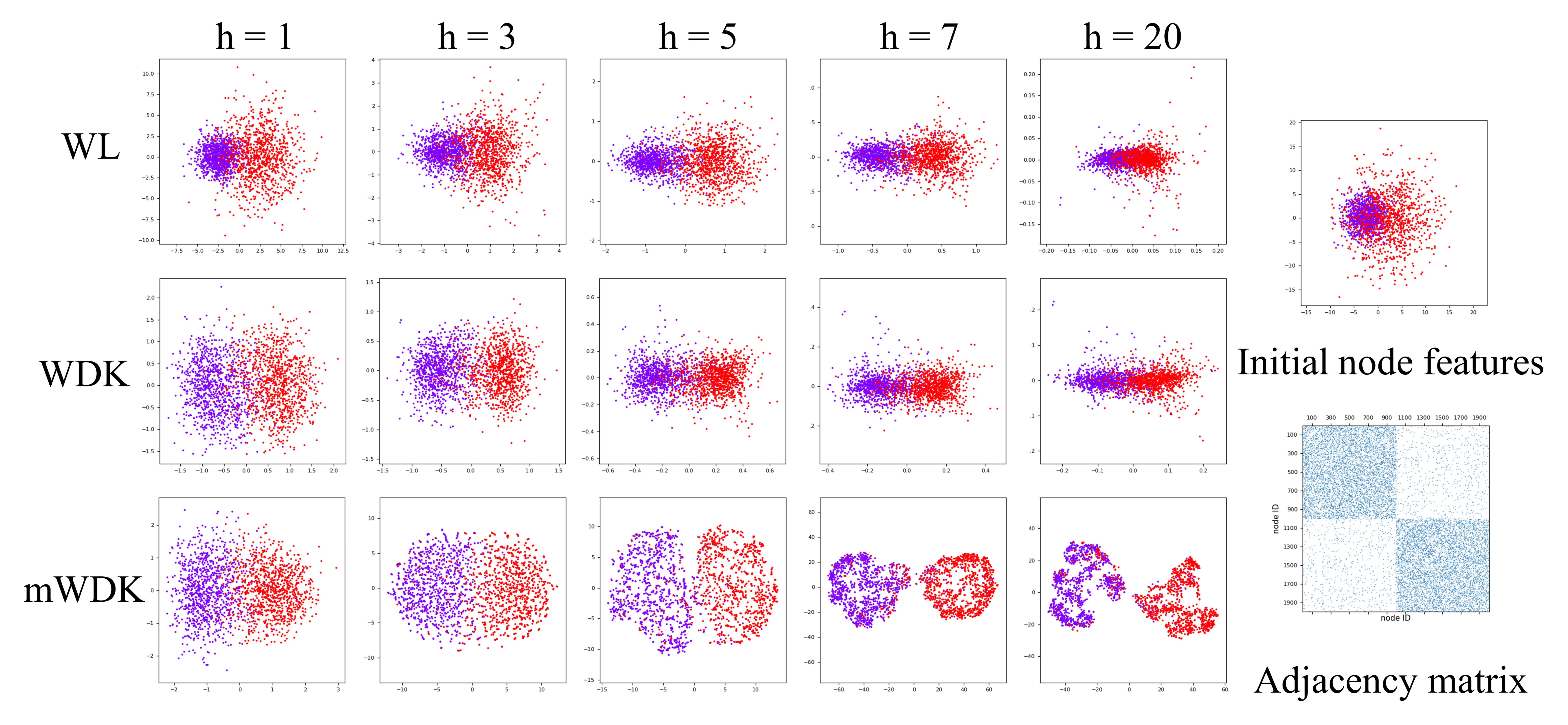}
  \vspace{-4mm}
	\caption{t-SNE's 2D visualization of the embedded spaces of WL, WDK, mWDK at different iterations $h$ on the same UE dataset used in Figure \ref{fig:sim}.}
	\label{fig:UE}
 \vspace{-4mm}
\end{figure}

Figure \ref{fig:UE} shows the visualizations of the embedded spaces  at different iterations of WL, WDK and mWDK.  
Note that the clusters have a significant overlap in the input space.


While WL could reduce the magnitude of density difference between the two clusters as $h$ increases, the two clusters are in fact harder to be separated because the overlapping region becomes denser (see the first row in Figure \ref{fig:UE}).
Further,
WDK has the same issue as WL in separating the two clusters, even though the two clusters achieve $S(\mathcal{C}_1) \approx S(\mathcal{C}_2)$ at $h \ge 1$. In contrast,
mWDK enables the separation of the two clusters at $h \in [5,20]$, while $S(\mathcal{C}_1) \approx S(\mathcal{C}_2)$ at $h \ge 1$.  



\begin{figure}[t]
    \centering
    \includegraphics[width=\linewidth]{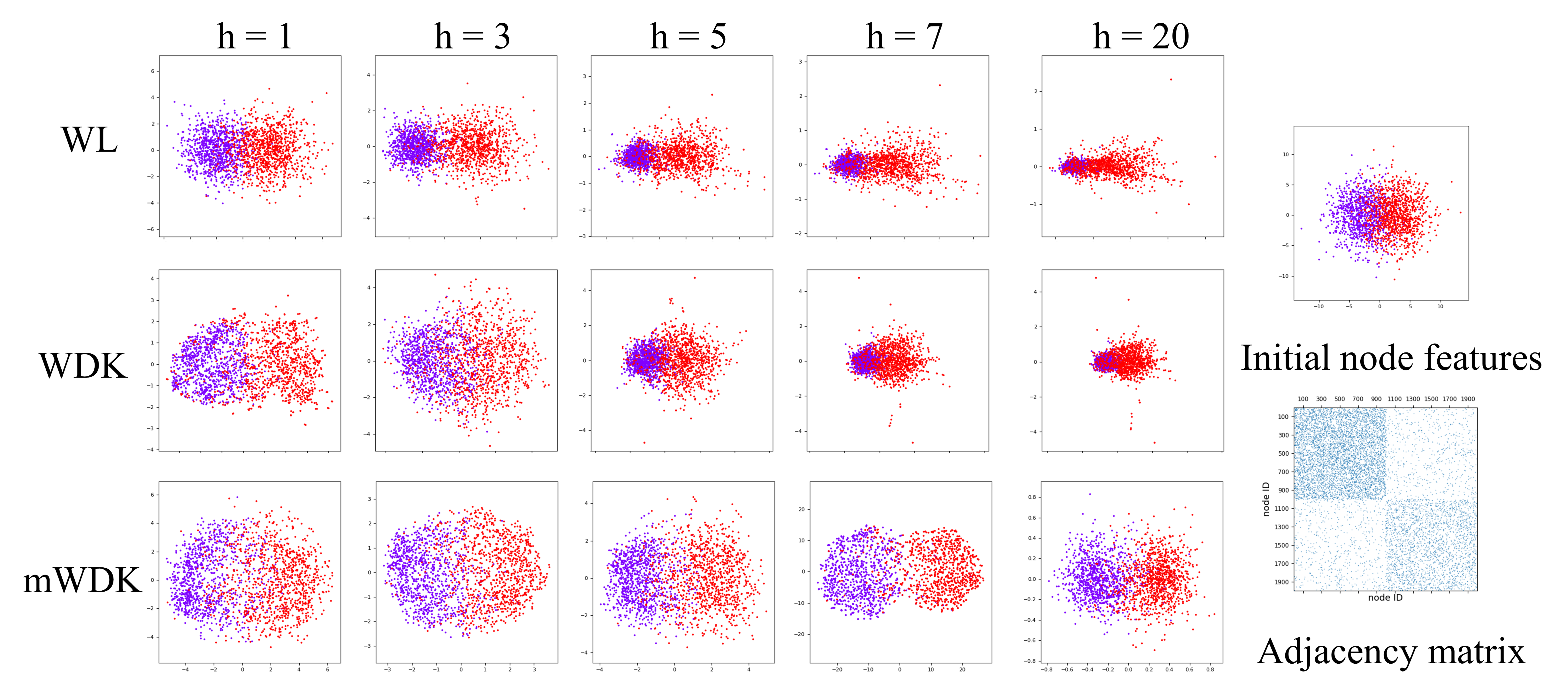}
     \vspace{-4mm}
    \caption{t-SNE's 2D visualization of the embedded spaces of WL, WDK, mWDK at different iterations $h$ on the EU dataset  having clusters with different node degrees (see Section \ref{Sec-Datasets}).}
    \label{fig:EU}
    \vspace{-3mm}
\end{figure}

\vspace{2mm}
\noindent
\textbf{Dataset having two clusters with equal node density and unequal node degrees}. These characteristics are shown in the two subfigures in the last column in
Figure \ref{fig:EU}. 

While WL has begun with equal node density  at $h=1$ for both clusters, the red cluster is sparser than the blue cluster at all values of $h>1$. This is due to the effect of imbalance node-degree distribution between the two clusters. WDK suffers from the same effect.
In contrast, mWDK managed to maintain approximately the same density for both clusters for all values of $h$. The separation between the two clusters could be best achieved at $h=7$. 

In summary, only mWDK satisfies the two conditions specified in Definition \ref{Def-Graph-Embedding} on the above two datasets.

\vspace{2mm}
\noindent
\textbf{The use of IK enables mWDK to produce clusters of equal density, preventing over-concentration}. 
Note that both WL and WDK in the above two examples, shown in Figures \ref{fig:UE} \& \ref{fig:EU}, exhibit the over-smoothing effect, i.e., as $h$ increases, all nodes in both clusters become increasingly concentrated. Yet, mWDK successfully resists the over-smoothing effect because of the repeated application of  Isolation Kernel (IK) in each iteration. IK has been previously shown to have a key ability to deal with clusters of varied densities \cite{ting2018isolation, ting2020-IDK}. 

Here we provide a reason of this ability via a property of Voronoi tessellation (revealed by a recent work \cite{VORONOICELLS}):
\begin{theorem}\label{thm:equal density}
    Let $\{\Lambda_i\}_{1}^{\psi}$ be a Voronoi tessellation of $\psi$ data points, where $\Lambda_i$ denotes the  Voronoi cell centered at $x_i$, and $M_X(\Lambda_i)$ denotes an estimator of the probability mass of $\Lambda_i$ based on a data set $X \subset \mathbb{R}^d$.
    The expectation $E[M_X(\Lambda_i)] = 1/\psi$ \cite{VORONOICELLS}.
\end{theorem}

In simple terms, each Voronoi cell in a Voronoi tessellation has approximately the same mass, as a result of large cells in sparse regions and small cells in dense regions.
Because IK employs multiple Voronoi tessellations \cite{DBSCAN-IK-AAAI2019,xu2021isolation} to estimate the similarity between any two points, Theorem \ref{thm:equal density} explains why the use of IK in each iteration of mWDK maintains clusters of equal mass, thus it 
prevents over-concentration. The inability to counter the effect of clusters of varied densities and over-concentration at $h>1$ in WDK is simply because only one application of IK is conducted.

\vspace{2mm}
\noindent
\textbf{The use of IK enables mWDK to maintain well-separated clusters at each iteration}; and
IK has the following property:

\begin{theorem}\label{thm:well-sep}
    Let $\delta= min_{x\in \mathcal{C}_1, y\in \mathcal{C}_2}\ \ell_{p}(\phi(x),\phi(y))$ be the inter-cluster distance of two clusters $\mathcal{C}_1$ and $\mathcal{C}_2$ in the IK-induced feature space $\phi$.
    IK $\kappa_\psi$ with $t$ Voronoi tessellations produces a maximum $\delta=\sqrt[p]{2t}$, if it holds that 
\begin{equation}\label{eq:ap}
\ell_{2}(z,z')\le \ell_{2}(x,y),
\end{equation}
for any $z, z'$ distinct points in either $\mathcal{C}_1$ or $\mathcal{C}_2$ and $x \in \mathcal{C}_1$, 
$y\in \mathcal{C}_2$.
\end{theorem}

In simple terms, Theorem \ref{thm:well-sep} shows that,  as long as each iteration of embedding in mWDK (line 4 in Algorithm \ref{alg:mWDK}) does not cause nodes from different clusters to get too close (i.e., $\delta$ still satisfies Eq (\ref{eq:ap})), then IK $\kappa_{\psi}$ applied at the next iteration of mWDK (line~3 in Algorithm \ref{alg:mWDK})  enhances the inter-cluster separation up to a maximum $\delta=\sqrt[p]{2t}$. This prevents over-smoothing with high $h$ from merging the two clusters, unlike that in WDK where IK is applied only once (line 1 in Algorithm~\ref{alg:WDK}). In contrast, the use of Gaussian Kernel in mWDK cannot maintain well-separated clusters at each iteration. The detailed discussion on Gaussian Kernel is provided in  Appendix.


In a nutshell, the over-smoothing has two distinct effects: \textbf{As $h$ increases, (a)  nodes in each cluster become increasingly concentrated, and (b) the effect of clusters of varied densities could not be prevented for every $h< \infty$. Effect (b) increases the likelihood of intermingling between different clusters as $h$ increases. IK applied at each iteration prevents both effects from taking hold.}

\section{Experimental settings}
\label{sec-experiments}

\subsection{Experimental setup and metric}
\label{append: setup}


mWDK has two parameters that need to be tuned: the number of iterations $\hslash $ and the number of sample points $\psi$ to produce IK. The optimal value of $\hslash$ is usually small and the specific values used in the experiments are shown in Table \ref{tab:ablation}. The values of $\psi$ used in the search are [2,4,6,8,16,32,64,128], but most datasets achieve the best result with $\psi=64$.

Each method is executed ten times, evaluated in terms of accuracy (ACC), normalized mutual information (NMI) and ARI \cite{estevez2009normalized}, and we report the average metrics for each competing method.

Here we present the datasets, existing graph embedding methods used in the assessment. The detailed experimental setup \& metric are provided in the Appendix. Our code is available \href{https://anonymous.4open.science/r/mWDK-code/}{\underline{here}}.

\subsection{Datasets}
\label{Sec-Datasets}
\noindent \textbf{Real World Datasets.} We choose datasets from citation networks (Cora, Citeseer, Pubmed \cite{sen2008collective}; DBLP and ACM \cite{wang2019heterogeneous}), social networks  (BlogCatalog \cite{LiHTL15}), product networks (AMAP and AMAC \cite{shchur2018pitfalls,tu2021siamese}) and a large-scale graph dataset Ogbn-products \cite{hu2020open}.

\noindent \textbf{Synthetic Datasets.} To better analyze the impact of imbalanced distribution on graph embedding and the distinctive characteristics of mWDK, we conducted several experiments on synthetic datasets. 
Four datasets with different characteristics are used:

\begin{itemize}
    \item ENode\_EDegree\_Easy (EEE): It has two clusters which have the same node density and the same average node degree, with a few inter-cluster edges. It serves as a control group for performance comparison.
    \item  ENode\_EDegree\_Hard (EEH): It has the same characteristics as EEE, except that it has more inter-cluster connections, posing an increased risk of over-smoothing for NAS methods.
    \item  UNode\_EDegree (UE): It has two clusters sharing identical average node degree, but one cluster exhibits a notably higher node density than the other, showcasing an imbalanced  distribution of nodes.
    \item  ENode\_UDegree (EU): It is designed to investigate the imbalanced distribution of node degrees. It has two clusters with equal node density but differing average node degrees. 

\end{itemize}


\subsection{Research questions and benchmarks}
The experiments are designed to answer two questions:
\begin{enumerate}
    \item Does the use of distributions of node attributes and node degrees have an impact on clustering outcomes?
    \item Does the optimization (as used in deep learning) improve the clustering outcome of that without optimization?
\end{enumerate}

To answer the first question, we compare WL, WDK and mWDK. Comparing WL and WDK allows us to examine the advantage of using the distribution of nodes only. Comparing WDK and mWDK investigates the advantage of using the distribution of node degrees. We also include (i) AGC \cite{zhang2023adaptive}, a WL-like scheme; (ii) NDLS \cite{zhang2021node}, a method considering the distribution of node degrees only;  and (iii) GSNN \cite{wijesinghe2022new}, a method which claims to have overcome the weaknesses of WL. AGC and GSNN do not consider any distribution at all.

To answer the second question, we include three SOTA  methods which employ optimization, i.e., PANE \cite{yang2020scaling}, a random walk method, DAEGC \cite{wang2019attributed}, an autoencoder method, and MAGI \cite{liu2024revisiting}, a contrastive learning method; as well as GAE \cite{kipf2016variational} and the versions of NDLS and GSNN which employ GAE to learn the weights. The GAE-based deep learning methods serve two purposes: they examine the advantage of (i) deep learning versions over optimization-free methods of NDLS and GSNN, and (ii) the improved versions of GAE using NDLS and GSNN over the base GAE. 

All these selected methods are representative graph embedding methods that excel in one of the four categories shown in Table \ref{tab:categories of embedding methods}.

\begin{table*}[!htp]
\vspace{-3mm}
\centering
\setlength{\tabcolsep}{2.5pt}
\caption{Comparing different graph embedding methods using the same Spectral Clustering (except DAEGC and MAGI which are end-to-end clustering methods) in NMI. OOM: out-of-memory error during the run \& OpF: optimization-free.}
\vspace{-3mm}
\begin{tabular}{l|rrrr|c|cccccccc|cccc}
    \toprule 
    \multirow{3}{*}{Dataset}       & \multicolumn{4}{c|}{Network characteristics} & \multirow{3}{*}{Metric} & 
    \multicolumn{8}{c|}{Deep learning and random-walk methods} & 
    \multicolumn{4}{c}{Optimization-free (OpF) methods} \\
    \cline{7-18}                            &                                              &                         &                                                            &                                              &     & 
    \multirow{2}{*}{GAE} & \multicolumn{2}{c}{NDLS} & \multicolumn{2}{c}{GSNN} & \multirow{2}{*}{PANE} & \multirow{2}{*}{DAEGC} & \multirow{2}{*}{MAGI} & \multirow{2}{*}{AGC} & \multirow{2}{*}{WL} & \multirow{2}{*}{WDK} & \multirow{2}{*}{mWDK} \\
    \cline{8-11}                            
    & $|\mathcal{V}|$ & $d$ & $|\mathcal{E}|$ & $k$ &  & 
     & GAE & OpF & GAE & OpF &  &  &  &  &  &  &  \\
    \midrule

    \multirow{3}{*}{EEE} 
        & \multirow{3}{*}{2,000} & \multirow{3}{*}{100} & \multirow{3}{*}{6,765} & \multirow{3}{*}{2} 
        & ACC & .938 & .976 & \underline{.991} & .977 & .985 & \underline{.991} & .984 & \textbf{.992} & .988 & .985 & .990 & \textbf{.992} \\
        & & & & & NMI & .675 & .860 & \textbf{.930} & .851 & .891 & \textbf{.930} & .891 & \textbf{.930} & .913 & .891 & \underline{.919} & \textbf{.930} \\
        & & & & & ARI & .767 & .916 & \underline{.964} & .910 & .939 & \underline{.964} & .935 & \textbf{.966} & .953 & .941 & .960 & \textbf{.966} \\
    \midrule

    \multirow{3}{*}{EEH} 
        & \multirow{3}{*}{2,000} & \multirow{3}{*}{100} & \multirow{3}{*}{7,991} & \multirow{3}{*}{2} 
        & ACC & .860 & .945 & .940 & .884 & .931 & .965 & .935 & \underline{.973} & .911 & .932 & .962 & \textbf{.986} \\
        & & & & & NMI & .416 & .699 & .682 & .485 & .655 & .785 & .664 & \underline{.792} & .577 & .660 & .774 & \textbf{.899} \\
        & & & & & ARI & .518 & .792 & .774 & .591 & .745 & .863 & .675 & \underline{.878} & .676 & .748 & .854 & \textbf{.943} \\
    \midrule

    \multirow{3}{*}{UE}  
        & \multirow{3}{*}{2,000} & \multirow{3}{*}{100} & \multirow{3}{*}{6,457} & \multirow{3}{*}{2} 
        & ACC & .822 & .851 & .879 & .844 & .862 & .889 & .880 & .863 & .859 & .852 & \textbf{.934} & \underline{.933} \\
        & & & & & NMI & .328 & .403 & .485 & .376 & .421 & .519 & .475 & .424 & .420 & .418 & \underline{.656} & \textbf{.668} \\
        & & & & & ARI & .413 & .493 & .574 & .473 & .522 & .605 & .577 & .527 & .515 & .494 & \textbf{.755} & \underline{.752} \\
    \midrule

    \multirow{3}{*}{EU}  
        & \multirow{3}{*}{2,000} & \multirow{3}{*}{100} & \multirow{3}{*}{6,337} & \multirow{3}{*}{2} 
        & ACC & .780 & .827 & .835 & .804 & .835 & .844 & \underline{.885} & .859 & .838 & .818 & .862 & \textbf{.948} \\
        & & & & & NMI & .284 & .398 & .433 & .351 & .415 & .446 & \underline{.502} & .421 & .408 & .412 & .485 & \textbf{.709} \\
        & & & & & ARI & .314 & .427 & .449 & \underline{.639} & .447 & .472 & .593 & .577 & .457 & .404 & .524 & \textbf{.803} \\
    \midrule

    \multirow{3}{*}{Cora} 
        & \multirow{3}{*}{2,708} & \multirow{3}{*}{1,433} & \multirow{3}{*}{5,429} & \multirow{3}{*}{7} 
        & ACC & .608 & .630 & .685 & .641 & .645 & .664 & .681 & \textbf{.760} & .689 & .634 & .696 & \underline{.748} \\
        & & & & & NMI & .410 & .468 & .537 & .486 & .489 & .508 & .524 & \textbf{.599} & .534 & .501 & .542 & \underline{.583} \\
        & & & & & ARI & .306 & .359 & .437 & .387 & .391 & .408 & .433 & \textbf{.576} & .447 & .381 & .467 & \underline{.530} \\
    \midrule

    \multirow{3}{*}{Citeseer} 
        & \multirow{3}{*}{3,327} & \multirow{3}{*}{3,703} & \multirow{3}{*}{4,732} & \multirow{3}{*}{6} 
        & ACC & .541 & .585 & .667 & .503 & \underline{.676} & .667 & .673 & \textbf{.693} & .670 & .667 & .663 & .672 \\
        & & & & & NMI & .276 & .362 & .417 & .287 & .403 & .415 & \underline{.419} & \textbf{.428} & .409 & .407 & .406 & .426 \\
        & & & & & ARI & .268 & .326 & .406 & \textbf{.474} & .428 & .409 & .425 & \underline{.445} & .421 & .416 & .401 & .411 \\
    \midrule

    \multirow{3}{*}{Wiki} 
        & \multirow{3}{*}{2,405} & \multirow{3}{*}{4,973} & \multirow{3}{*}{17,981} & \multirow{3}{*}{17} 
        & ACC & .303 & .364 & .470 & .352 & .513 & \textbf{.535} & \underline{.515} & .468 & .479 & .485 & .460 & .512 \\
        & & & & & NMI & .225 & .308 & .455 & .336 & .470 & \textbf{.554} & \underline{.511} & .482 & .459 & .463 & .479 & .502 \\
        & & & & & ARI & .120 & .159 & .285 & .177 & .289 & \textbf{.358} & .259 & .293 & .147 & .262 & .265 & \underline{.343} \\
    \midrule

    \multirow{3}{*}{DBLP} 
        & \multirow{3}{*}{4,057} & \multirow{3}{*}{334} & \multirow{3}{*}{3,528} & \multirow{3}{*}{4} 
        & ACC & .649 & .664 & .629 & .667 & .675 & \underline{.766} & .505 & .764 & .676 & .645 & .618 & \textbf{.816} \\
        & & & & & NMI & .356 & .391 & .355 & .398 & .406 & \underline{.487} & .235 & .473 & .388 & .354 & .395 & \textbf{.549} \\
        & & & & & ARI & .296 & .355 & .308 & .361 & .373 & \underline{.471} & .122 & .479 & .325 & .271 & .370 & \textbf{.577} \\
    \midrule

    \multirow{3}{*}{ACM} 
        & \multirow{3}{*}{3,025} & \multirow{3}{*}{1,870} & \multirow{3}{*}{13,128} & \multirow{3}{*}{3} 
        & ACC & .826 & .876 & .893 & .889 & .802 & \underline{.893} & .875 & .891 & .836 & .837 & .858 & \textbf{.912} \\
        & & & & & NMI & .535 & .637 & .678 & .658 & .503 & \underline{.695} & .628 & .661 & .551 & .556 & .578 & \textbf{.707} \\
        & & & & & ARI & .560 & .680 & .702 & .701 & .519 & \underline{.729} & .663 & .709 & .577 & .586 & .624 & \textbf{.757} \\
    \midrule

    \multirow{3}{*}{AMAP} 
        & \multirow{3}{*}{7,650} & \multirow{3}{*}{745} & \multirow{3}{*}{119,081} & \multirow{3}{*}{8} 
        & ACC & .474 & .482 & .688 & .441 & .672 & .629 & .667 & \underline{.778} & .693 & .663 & .739 & \textbf{.785} \\
        & & & & & NMI & .251 & .275 & .620 & .303 & .533 & .482 & .674 & \underline{.690} & .637 & .656 & .664 & \textbf{.717} \\
        & & & & & ARI & .163 & .171 & .521 & .203 & .471 & .398 & .503 & \textbf{.805} & .528 & .497 & .573 & \underline{.611} \\
    \midrule

    \multirow{3}{*}{AMAC} 
        & \multirow{3}{*}{13,752} & \multirow{3}{*}{767} & \multirow{3}{*}{574,418} & \multirow{3}{*}{10} 
        & ACC & .291 & .265 & .432 & .304 & .475 & .452 & .430 & \underline{.537} & .502 & .454 & .485 & \textbf{.547} \\
        & & & & & NMI & .182 & .185 & .415 & .208 & .400 & .408 & .441 & \underline{.479} & .455 & .421 & .436 & \textbf{.481} \\
        & & & & & ARI & .223 & .114 & .269 & .230 & .269 & .264 & .259 & \underline{.376} & .251 & .248 & .284 & \textbf{.393} \\
    \midrule

    \multirow{3}{*}{Pubmed} 
        & \multirow{3}{*}{19,717} & \multirow{3}{*}{500} & \multirow{3}{*}{44,338} & \multirow{3}{*}{3} 
        & ACC & .638 & .642 & .676 & \underline{.698} & .638 & .635 & .636 & .633 & .618 & .621 & .618 & \textbf{.699} \\
        & & & & & NMI & .224 & .269 & .309 & .291 & .275 & .311 & .301 & .306 & .321 & .316 & .320 & \textbf{.335} \\
        & & & & & ARI & .241 & .246 & .308 & .309 & .255 & \underline{.313} & .287 & .207 & .284 & .281 & .292 & \textbf{.325} \\
    \midrule

    \multirow{3}{*}{Blogcatalog} 
        & \multirow{3}{*}{5,169} & \multirow{3}{*}{8,189} & \multirow{3}{*}{171,743} & \multirow{3}{*}{6} 
        & ACC & .406 & .404 & .412 & .426 & .471 & .503 & .515 & \underline{.798} & .519 & .570 & .667 & \textbf{.849} \\
        & & & & & NMI & .215 & .266 & .247 & .235 & .304 & .404 & .465 & \underline{.713} & .407 & .439 & .522 & \textbf{.678} \\
        & & & & & ARI & .157 & .201 & .139 & .157 & .252 & .253 & .340 & \underline{.633} & .357 & .361 & .467 & \textbf{.682} \\
    \midrule   
    \multirow{3}{*}{Ogbn-products} 
        & \multirow{3}{*}{2,449,029} & \multirow{3}{*}{100} & \multirow{3}{*}{61,859,140} & \multirow{3}{*}{47} 
        & ACC & \multirow{3}{*}{OOM} & \multirow{3}{*}{OOM} & .369 & \multirow{3}{*}{OOM} & \multirow{3}{*}{OOM} & .398 & \multicolumn{1}{c}{\multirow{3}{*}{OOM}} & \underline{.428} & .381 & .388 & .412 & \textbf{.435} \\
        & & & & & NMI &                      &                      & .497 &                      &                      & .515 & \multicolumn{1}{c}{} & .557 & .503 & .505 & \underline{.535} & \textbf{.583} \\
        & & & & & ARI &                      &                      & .216 &                      &                      & .220 & \multicolumn{1}{c}{} & .219 & .208 & .212 & \underline{.241} & \textbf{.252} \\
    \bottomrule
\end{tabular}
\label{tab:result}
\end{table*}

\section{Experiments}
\label{sec-results}

\label{sec-clustering-outcomes}

We evaluate clustering performance across four synthetic and ten real-world datasets using three commonly used metrics: Accuracy (ACC), Normalized Mutual Information (NMI), and Adjusted Rand Index (ARI). Our analysis focuses on NMI, with ACC and ARI provided as complementary measures. 

The NMI results in Table  \ref{tab:result} provide the answers for each of the two research questions:


\noindent\textbf{Answers to RQ1: (1a) Standard NAS-based methods fail to address imbalanced distributions and heterophilic connectivity.} WL, AGC, and GSNN all employ the neighborhood aggregation strategy (NAS) without  modeling of node and degree distributions. Consequently, they achieve strong performance on homophilic graphs (e.g., EEE) but suffer severe performance degradation on: (i) graphs with imbalanced node or degree distribution (e.g., UE, EU), where the imbalanced aggregation treats dense and sparse regions equally, causing imbalanced smoothing  rates that erode node distinguishability; and (ii) graphs with strong heterophily (e.g., EEH, real-world networks), where aggregating dissimilar neighbors exacerbates over-smoothing.
    
\noindent\textbf{(1b) Modeling node distribution enhances separability under varying densities.} WDK consistently outperforms WL across all datasets, with the largest significant improvement observed on UE, where the two clusters have a huge difference in node densities. This highlights the benefit of incorporating node distribution. 
    
\noindent\textbf{(1c) Modeling degree distribution mitigates imbalanced smoothing.} mWDK further improves upon WDK on all datasets, most notably on EU (NMI: 0.71 vs. 0.49), where the two clusters have a huge difference in node degrees. This confirms that explicitly accounting for degree distribution effectively addresses imbalanced smoothing rates, a key factor exacerbating over-smoothing in NAS-based methods.
    
\noindent\textbf{(1d) Joint modeling ensures consistent superiority.} (i) WDK (node-distribution aware only) demonstrates clear advantages on UE, and (ii) NDLS (degree-distribution aware only) on EU. But, both methods show subdued NMI on the real-world graphs (avg. NMI: WDK: 0.48, NDLS: 0.36). This reveals a critical insight: node density and degree imbalance induce distinct effects, and neither can be fully resolved without joint modeling. In contrast, mWDK integrates both distributions within a unified kernel framework, addressing the adverse effects of over-smoothing, specifically the loss of node distinguishability (avg. NMI: 0.55).


\noindent\textbf{Answers to RQ2: (2a) Optimization-free methods are competitive with or better than deep learning methods.} GAE performs poorly across all datasets. While integrating structural priors such as GSNN and NDLS improves its performance (this result is consistent with the findings in the graph classification context \cite{wijesinghe2022new}), the overall performance remains subdued. Among the deep learning methods, DAEGC and MAGI, which are specifically designed for graph clustering, achieve competitive results. Yet, mWDK matches or exceeds them without any optimization. This demonstrates that the explicit distribution-aware modeling without optimization can produce embeddings of equal or higher quality than those with deep learning. 

\noindent\textbf{(2b) No-optimization mWDK has superior scalability on large-scale graphs.} On the Ogbn-products dataset (having 2.4M nodes and 61M edges), mWDK achieves the best result among all the methods  which can complete the task. The closest contender is the deep learning MAGI, but it still performs worse than mWDK. 


Overall, mWDK achieves the highest NMI on 12 out of 14 datasets (having equal NMI on 1 dataset and slightly weaker on 2, compared with the closest contender MAGI). These results establish three key conclusions:  
(1) Distributional information, including both node vectors and node degrees, is essential for effective graph representation and clustering performance;  
(2) The Joint modeling of both types of distributions leads to improved performance on real-world networks;  
(3) The optimization-free mWDK provides an efficient and scalable solution for unsupervised community detection.

\section{Time complexity and Scale-up test}
 For a graph with $n$ nodes and $m$ edges, the time complexity of a single IK mapping is $O(\psi tn)$. The resulting embedding has a dimensionality of $\psi t$ . The time complexity of a single-iteration of WL is $m\psi t$ . Thus,  the overall time complexity of the mWDK is $O(\psi th(n+m))$, which is linear to $n$ and $m$.

It is interesting to note that mWDK usually achieves better clustering outcomes than WDK with a fewer number of  iterations, sometimes the difference is huge. The $h$ settings used are shown in the last three columns in Table \ref{tab:ablation}. The huge difference in the $h$ setting can be found on the Citeseer and Pubmed datasets. As a result, mWDK can run significantly faster than WDK.

\begin{figure}[ht]
    \centering
    \includegraphics[width=.93\linewidth]{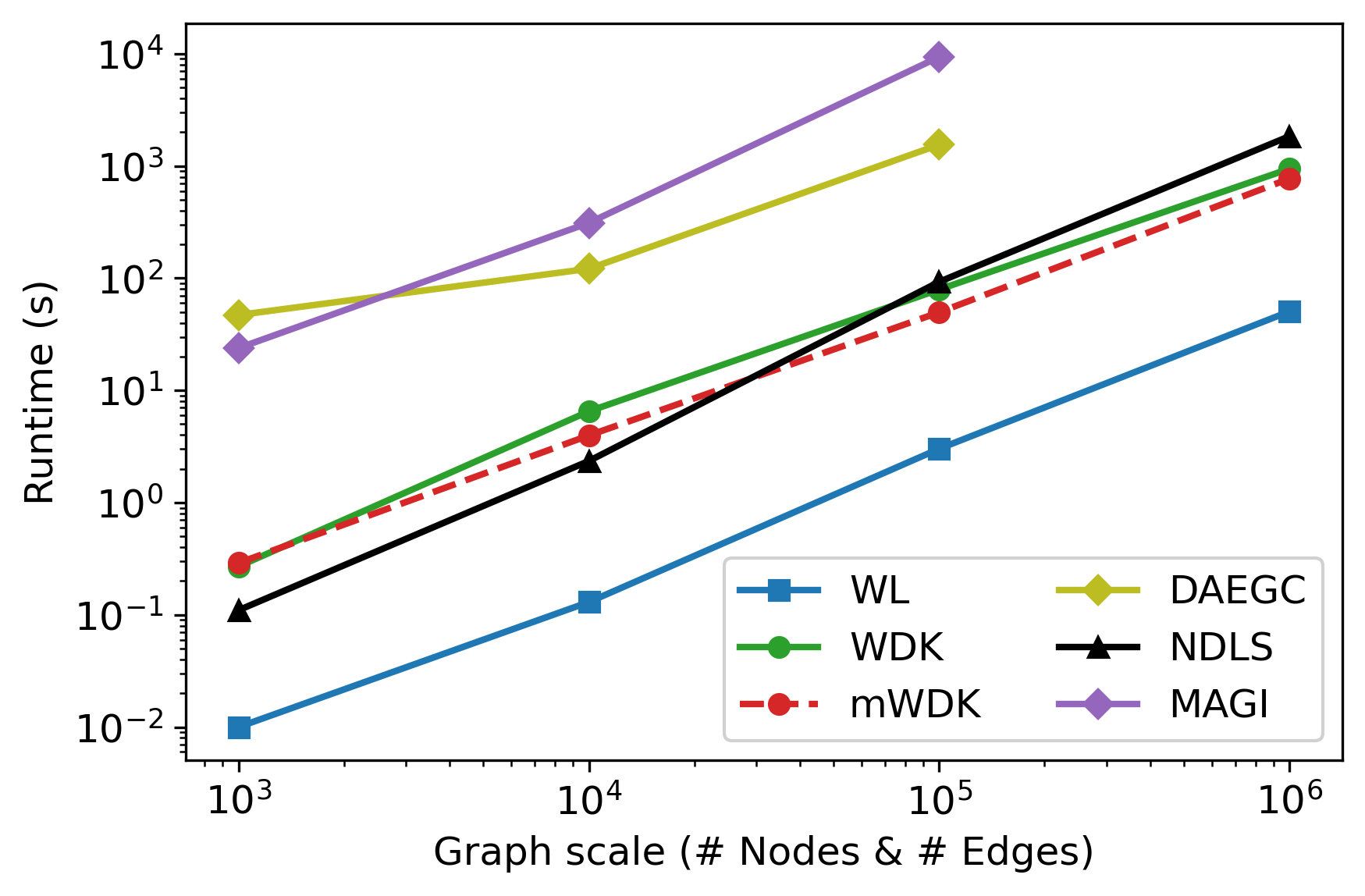}
    \caption{Scaleup test results of five embedding methods.}
    \label{fig:scale-up}
\end{figure}

Figure \ref{fig:scale-up} shows the scaleup test results comparing five embedding methods\footnote{
 For DAEGC and MAGI, we used the default parameters specified in their paper \cite{wang2019attributed,liu2024revisiting}. For all other methods, we used the best parameters for clustering task. Only the graph embedding runtimes are reported.}. mwDK, WDK and NDLS have comparable runtimes. Consistent with our previous analysis, mWDK does not require more time than WDK due to its reduced number of iterations compared to WDK. DAEGC has the longest runtime and cannot be tested on larger datasets due to its large memory requirement.
 \section{Ablation studies}
We conduct three ablation studies in this section.

\textbf{First ablation study:} We compare the relative performance of the data dependent kernel and data independent kernel\footnote{Recall that WDK relies on a base kernel  $\kappa(u,v) = \left< \phi(u), \phi(v) \right>$. 
A recent work has shown that Isolation Kernel (IK) \cite{ting2018isolation} is better than Gaussian kernel (GK) in kernel mean embedding when used for point and group anomaly detections \cite{ting2020-IDK,IDK-TKDE2023}. Our result in Table \ref{tab:ablation} is consistent with this result.}, i.e.,  IK and Gaussian kernel (GK) on WDK. As shown in columns \#2 \& \#3 of Table \ref{tab:ablation}, IK outperforms GK on all datasets, without exception. The largest NMI difference occurred on the UE dataset. This is because IK can handle clusters of varied densities a lot better than GK. The first ablation study on WDK is shown in columns \#2  \&  \#3: GK versus IK. The second ablation study on mWDK is shown in columns \#4 to \#7:  three normalization methods and whether a concatenation of feature sets of all iterations is used.  The third ablation study on mWDK is shown in columns \#7 to \#10, where `all' combines three aggregation functions: min, max and average. The average aggregation is shown in column \#7:`wl,concat\ding{55}').
Because of high computational cost of running wl, concat\ding{51},
the largest dataset Ogbn-products is not used in this experiment.

In addition, we had replaced IK in mWDK with GK; but GK was too sensitive to the parameters at each iteration, and it failed to produce any good results (therefore, the results are omitted).

\begin{table*}[!htp]
\caption{The NMI results of three ablation studies and the best parameter $h$ used in WL, WDK and mWDK (in the last three columns). 
}
  \label{tab:ablation}
\resizebox{\textwidth}{!}{
\begin{threeparttable}
\begin{tabular}{l|cc|cccc|ccc|rrr}
\toprule
\multirow{3}{*}{Dataset}& \multicolumn{9}{c|}{NMI} & \multicolumn{3}{c}{Parameter $h$}    \\\cline{2-13}
 & \multicolumn{2}{c|}{WDK} & \multicolumn{7}{c|}{mWDK} & \multirow{2}{*}{WL} & \multirow{2}{*}{WDK} & \multirow{2}{*}{mWDK}\\
& GK & IK & sym,concat\ding{55} & rw,concat\ding{55}  & wl,concat\ding{51} & wl,concat\ding{55} &min&max & all \\ \hline
EEE         & .919          & .919          & .913        & .913      & .930    & .930  &.102&.879      & .930         & 12           & 5            & 3            \\
EEH         & .685        & .774        & .856        & .868      & .899   & .899      &.133&.754    & .899        & 16           & 6            & 2            \\
UE          & .435        & .656        & .685         & .656      & .689 & .668       &.138&.446    & .582              & 8            & 3            & 2            \\
EU          & .414         & .485         & .641        & .683    & .723    & .709    &.029& .457 & .651             & 7          & 2            & 4       \\ \hline
Cora        & .512        & .542        & .568        & .567       & .567  & .587  &.041&  .505   &     .557       & 10           & 7            & 3            \\
Citeseer    & .417        & .423       & .447        & .447        & .425   & .429   &.112&.409   &   .436           & 22           & 28           & 2            \\
Wiki        & .476        & .495        & .497        & .477     & .479   & .504   &.147&  .438 &   .441        & 3            & 1            & 1            \\
DBLP        & .341        & .367        & .486        & .499      & .549  & .543    &.254& .370 &     .420         & 2            & 1            & 5            \\
ACM         & .565        & .595        & .711        & .680      & .713 & .710     &.117& .622 &     .636        & 4            & 2            & 3            \\
AMAP        & .681        & .685        & .655        & .661       & .726  & .717     &.315&.578 &     .625            & 5            & 6            & 3            \\
AMAC        & .401        & .413        & .388        & .462       & .472  & .466    &.206&.361  &     .203         & 2            & 1            & 3            \\
Pubmed      & .301        & .332        & .252        & .311       & .313 & .335     &.136&.330   & .323          & 24           & 37           & 15           \\
Blogcatalog & .484          & .522          & .372        & .384       & .697   & .687 &.038&.237  &  .282            & 2            & 2            & 2            
     \\ \bottomrule
\end{tabular}
  \end{threeparttable}
  }
\end{table*} 




\textbf{Second ablation study:} We ablate two design choices in mWDK: (i) the normalization used in Eq. (7), and (ii) the strategy for constructing the final embedding.

For (i), we conducted an ablation study on mWDK wrt  the choice of normalization in Eq \ref{eqn-weighted-distribution}. Specifically, we considered three common graph normalizations:
\begin{itemize}
    \item Symmetric  Normalization: $\mathbf{W}_{\text{sym}}=\mathbf {D}^{-1/2}   \mathbf {A}\mathbf {D}^{-1/2}$
    \item Random Walk Normalization: $\mathbf{W}_{\text{rw}}=\mathbf {A}\mathbf {D}^{-1}   $
\item WL Normalization: $\mathbf{W}_{\text{wl}}=\mathbf {D}^{-1}   \mathbf {A}$
\end{itemize}
where $\mathbf{A}$ and $\mathbf{D}$ denote the adjacency matrix and degree matrix of $\mathcal{G}$, respectively.
Note that GCN-based models use symmetric 
matrices.
The other two matrices are not symmetric\footnote{The matrix induced by WL is sometimes called “random walk regularization” in the literature, but this is misleading. In fact, the random walk matrix and the WL-induced matrix are transposes of each other. They have similar forms, but different interpretations. For instance, in WL, all neighboring nodes have equal weights. In random walk, nodes with higher degrees have lower weights. The choice of the matrix depends on the task and data characteristics.}. 

 \begin{table*}[t]
    \centering
    \caption{NMI of WL, WDK, mWDK on the Cora dataset
    under different settings of noise edges.} 
    \label{tab: nmi noise}
    \begin{tabular}{cccc}
        \toprule
        WL & WDK & mWDK \\ 
        \midrule
        
        \begin{minipage}[c]{.3\textwidth}
            \centering
            \raisebox{-.4\height}{
                \includegraphics[width=\linewidth]{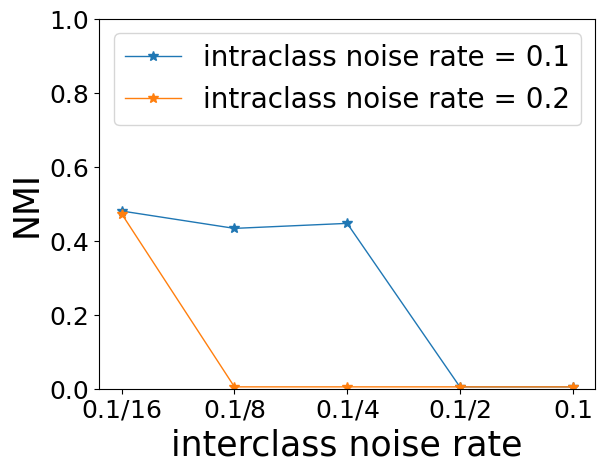}
            }
        \end{minipage}
        & \begin{minipage}[c]{.3\textwidth}
            \centering
            \raisebox{-.4\height}{
                \includegraphics[width=\linewidth]{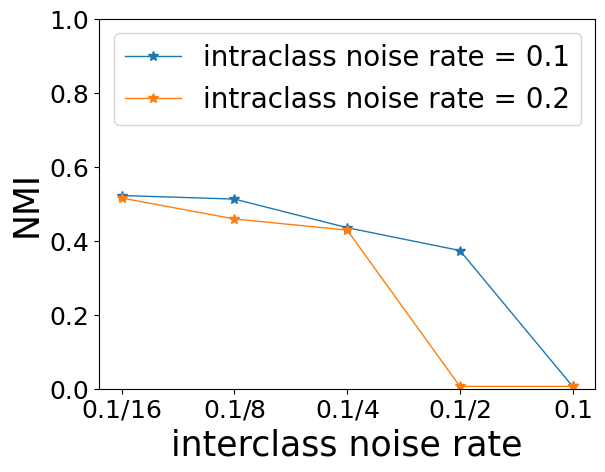}
            }
        \end{minipage}
        & \begin{minipage}[c]{.3\textwidth}
            \centering
            \raisebox{-.4\height}{
                \includegraphics[width=\linewidth]{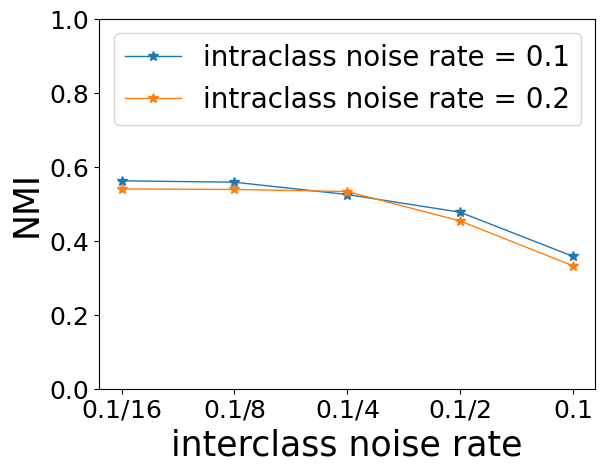}
            }
        \end{minipage} \\
        \bottomrule
    \end{tabular}
\end{table*}

For (ii), we compare the final feature set of the embedding.
The original WL scheme \cite{togninalli2019wasserstein}   concatenates the feature set of every iteration to form the final feature set for embedding. This enables the final embedding to capture information from all iterations. However, as the number of dimensions grows with the number of iterations, its use increases the computational cost significantly, which is $\psi th(n+mh)$. Because of the high runtime cost, we have used the feature set of the last iteration only in our implementation.


The result of using the original WL scheme with concatenation is included in the `wl,concat\ding{51}' column in  Table~\ref{tab:ablation}. The results of the second ablation study is summarized as follows:
\begin{itemize}
    \item The WL normalization performs the best in most datasets among the three normalization methods. A huge difference is on the Blogcatalog dataset.
    \item There is no clear advantage of using a concatenation of feature sets from all iterations over using the feature set from the last iteration only. A pair-wise Wilcoxon signed-rank test \cite{1944Individual} shows that there is no significant difference between them at  0.05 level.  
 
\end{itemize}

\textbf{Third ablation study:}
Our work so far has focused on a single aggregation function in NAS, that is, average.
A recent work suggests combining multiple aggregators to enhance the performance of a single aggregation function \cite{Aggregation-NIPS2020}.

Following this work, we examine a combination of three aggregation functions: minimum, maximum and average. We present the results comparing individual functions against their combination (indicated as the `all' column) in Table \ref{tab:ablation}. Our result shows that the best result is obtained by the average function (shown in the `wl,concat\ding{55}' column), and the combined function does not perform better than the average.

Two factors could have contributed to this outcome. First, the original assessment was conducted in the supervised regression and classification context \cite{Aggregation-NIPS2020}. The unsupervised clustering task studied here provides a different challenge. Second, the distributional information was not employed in their assessment \cite{Aggregation-NIPS2020}.

\section{Effect of noise edges}
We assess the performance of WL, WDK, and mWDK across various settings of noise edges. In a graph, noise edges are introduced by adding connections between nodes of different clusters (interclass noise) and removing connections between nodes within the same cluster (intraclass noise).

The results in Table~\ref{tab: nmi noise} show that
WL's NMI declines as the number of interclass noise edges increases. In contrast, WDK and mWDK maintain rather stable results across all noise edge settings, with mWDK achieving the highest NMI. The intraclass noise edges have almost no effect on mWDK for two reasons. First, fewer edges between the nodes within one cluster lead to  a sparser cluster, and IK is known to be adaptive to clusters of varied densities. According to Theorem \ref{thm:well-sep}, IK can maintain well-separated
clusters at each iteration, regardless of the relative density between clusters, as long as the connections within each cluster can satisfy the condition stated in Equation  \ref{eq:ap}. Second, fewer nodes in a cluster often have a minimum impact on the connectivity within a cluster, as long as the change is not so drastic as to alter the connectivity of being a cluster. 
\section{Sensitivity}

\begin{figure}[!htp]
        \centering
        \includegraphics[width=.95\linewidth]{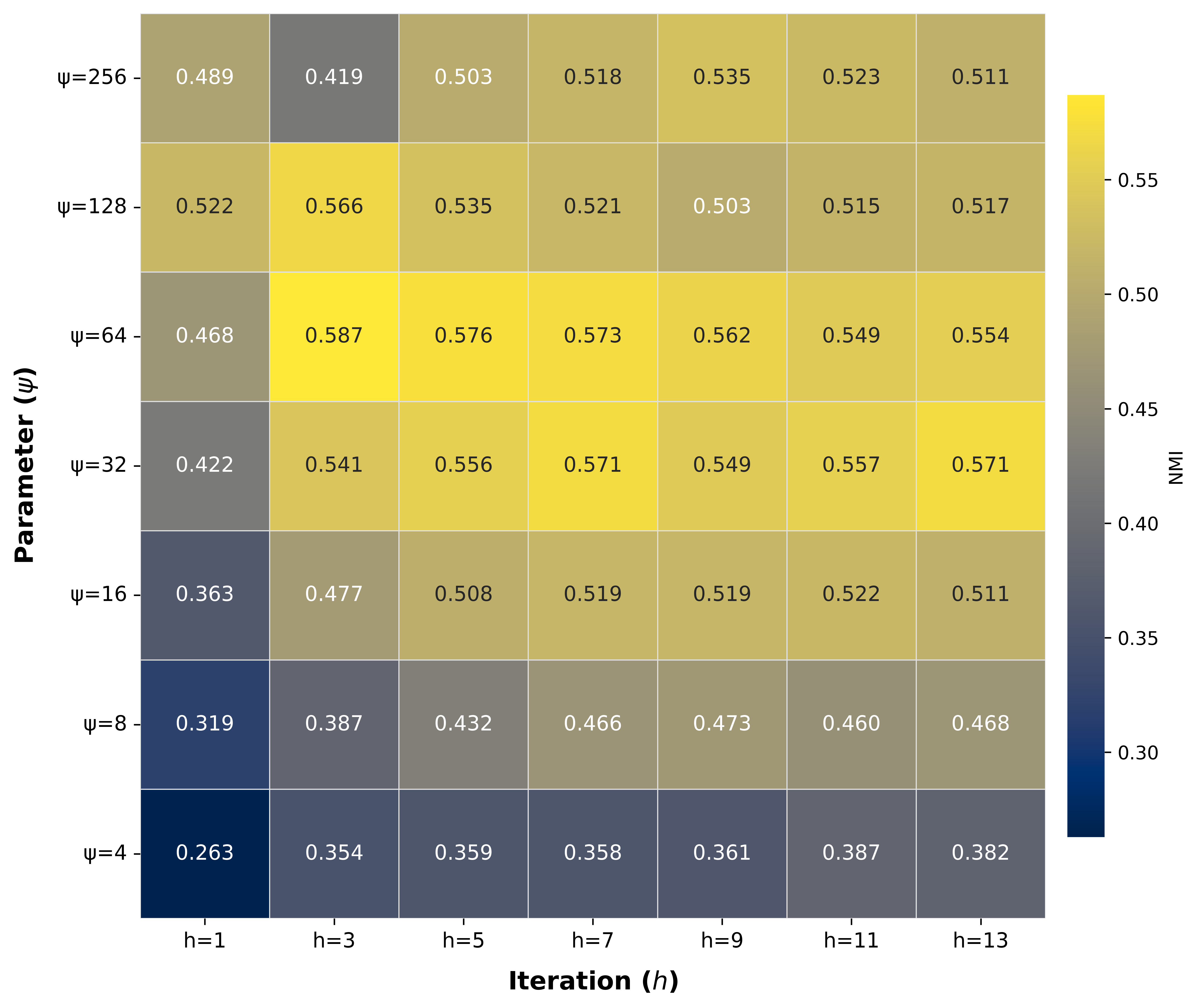}
        \caption{ mWDK's sensitivity to parameters $\psi$ and $h$ on Cora.}
        \label{fig:sensitivity}
    \end{figure}

Figure \ref{fig:sensitivity} shows an example result examining mWDK's sensitivity with respect to the parameters $\psi$ and $t$. It indicates that mWDK is is robust to these two parameters in the ranges $\psi \in [32,64,128]$ \& $h \in [3,5,7]$. 

\section{Significance rank test}
\label{appendix: srt}
\begin{figure}[h]
	\centering
	\includegraphics[width=.85\linewidth]{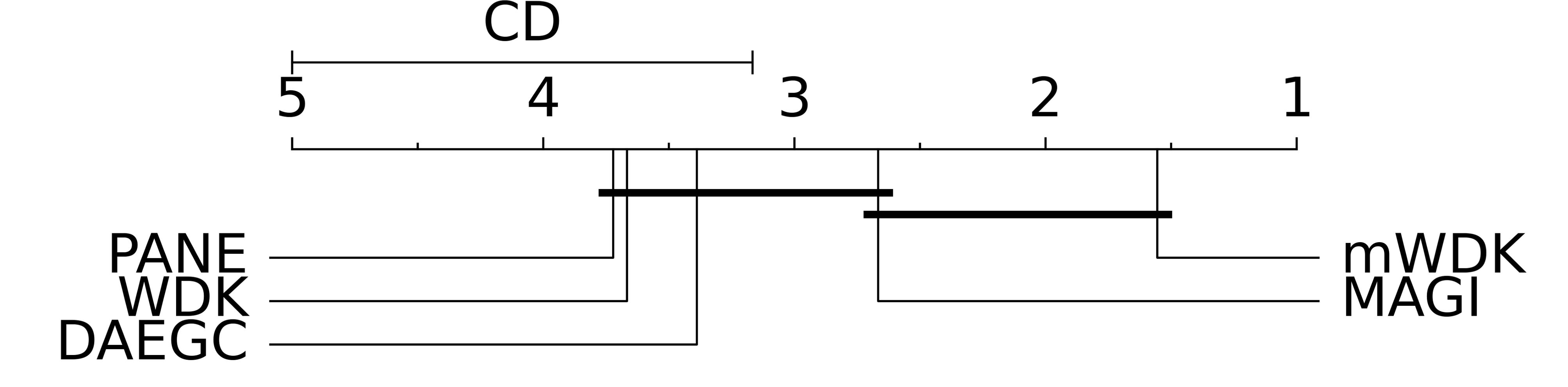}
	\caption{Friedman-Nemenyi test result for the top five embedding methods (shown in Table \ref{tab:result}) on NMI at significance level 0.1.}
	\label{fig:Sign-test}
\end{figure}

The significance rank test shown in Figure \ref{fig:Sign-test} reveals that mWDK is significantly better than the key contenders: DAEGC, WDK and PANE, though not significant with the second ranked MAGI. Notably, MAGI, DAEGC and PANE are not significantly different from the no-optimization WDK.   This is consistent with the results of two recent works using SGC \cite{wu2019simplifying} and AGC \cite{zhang2023adaptive}, without optimization, where both have comparable performance to deep learning methods. The no-optimization mWDK  further strengthens these findings. 

\section{Conclusions}


The proposed graph embedding method, Weighted Distributional Kernel mWDK introduces three novel contributions: First, it is the pioneering method to jointly incorporate node attribute and degree distributions during aggregation. Second, it operates without the need for optimization. Finally, by integrating this distribution-aware design into WL, it effectively neutralizes the over-smoothing effects in NAS-based methods in networks characterized by these dual distributions.



Our study delivers three important messages.
First, distributional information is critical to community detection but often overlooked, leading to suboptimal embeddings and degrading clustering performance.
Second, the adverse effects of oversmoothing can be mitigated by incorporating distributional information, specifically by replacing data-independent aggregation in Neighborhood Aggregation Strategies (NAS) with a data-dependent kernel such as the Isolation Kernel.
Third, the joint modeling of both distributions of nodes and node degrees is essential; modeling one type of distribution alone is insufficient for robust performance. mWDK with the joint modeling achieves significant performance gains over WL and WDK that model no or one type of distribution only.

Experiments show that optimization-free mWDK matches or exceeds state-of-the-art deep learning methods, scales efficiently to large networks like Ogbn-products, and remains robust to edge perturbations. 


 
%

\bibliographystyle{IEEEtran}
\bibliography{references}

{\appendix

\section{Dataset details}
\subsection{Intuition}
\begin{figure}[h]
    \centering
    \includegraphics[width=\linewidth]{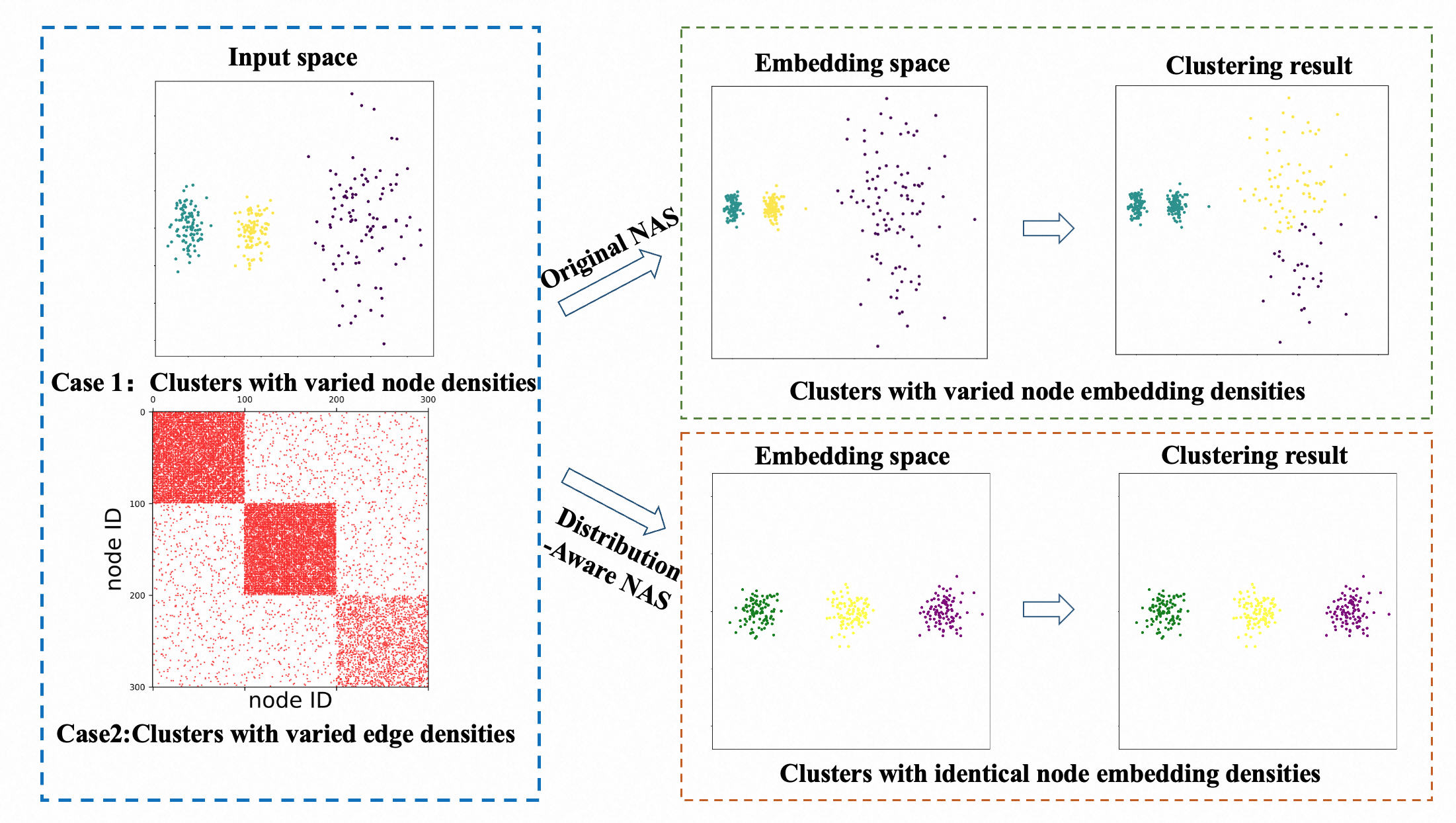}
    \caption{(A) The distribution of the varied densities of clusters
of node vectors and the distribution of varied edge degrees
in input space. (B) Any one of these two cases leads to clusters of embedded points with varied densities and an incorrect clustering result in the current NAS embedding. (C) Once the two distributions are considered, these issues are addressed to produce the desired clustering result.}
\end{figure}
\subsection{Real world datasets}
The details of these datasets are described below:
\begin{itemize}
    \item Cora, Citeseer and Pubmed are three widely used citation network benchmark datasets that consist of scientific papers from different domains and their citation relationships. Each node is a paper which labeled with a category and encoded with a binary or TF/IDF word vector \cite{sen2008collective}.
    \item WIKI is a word co-occurrence network collected from English Wikipedia pages, each node is a webpage  represented by a vector with 4973 dimensions.
    \item DBLP and ACM \cite{wang2019heterogeneous} are networks of authors and papers, respectively. DBLP nodes are authors with bag-of-words keyword features. ACM nodes are papers from KDD, SIGMOD, SIGCOMM, MobiCOMM, and VLDB, with 3 categories.
    \item AMAP and AMAC \cite{shchur2018pitfalls,tu2021siamese} are extracted from Amazon’s co-purchase network, where nodes represent products, edges represent frequent co-purchases, features represent encoded reviews, and labels are predefined product categories.
    \item  BlogCatalog \cite{LiHTL15} is a blog of a social network, which consists of nodes representing the bloggers and edges indicating the interactions between them.
     \item Ogbn-products \cite{hu2020open}  is  a large-scale graph dataset having more than 2 million nodes and 61 million edges. Nodes represent products sold on Amazon. Edges between two products indicate that they are frequently purchased together.
\end{itemize}
\subsection{Synthetic Datasets}
Specifically, these datasets were generated in the following way, controlled by the parameters specified in Table \ref{tab:synthetic-datasets}:
\begin{enumerate}
	\item   They all have two clusters in which nodes are drawn from two 100-dimensional Gaussian Distributions $N(\mu_1,\sigma_1)$ and $N(\mu_2,\sigma_2)$, respectively. 
	\item   The nodes in the same cluster  are connected with a probability $\alpha$; while the nodes from different clusters are connected with a lower probability $\beta < \alpha$. 
    The higher $\beta$ is, the more difficult to separate the two clusters.
	
	\item  The distributions of nodes and node degrees  are controlled by $\sigma$ and $\alpha$, respectively. For instance, to make $\mathcal{C}_1$ sparser than $\mathcal{C}_2$ on the UE dataset, $\sigma_1 > \sigma_2$  and the two clusters with same average degree by setting $\alpha_1 = \alpha_2$.
\end{enumerate}
\begin{table}[!htp]
	\begin{threeparttable}	
		   \caption{Parameters used to generate the synthetic datasets. $N(\mu,\sigma)$ denotes a single-dimensional Gaussian distribution; and the same $\mu,\sigma$ are used to generate the 100-dimensional Gaussian distribution.}

			\begin{tabular}{lcccc}
				\toprule
				
				Dataset                & $(\mu_1,\mu_2)$ & $(\sigma_1,\sigma_2)$ & $(\alpha_1,\alpha_2)\times 10^{-3}$ & $\beta\times 10^{-4}$ \\
				\midrule
				
    			EEE & (0,5)           & (10,10)               & (6,6)                               & 6                     \\
				EEH & (0,5)           & (10,10)               & (6,6)                               & 18                    \\
				UE       & (0,5)           & (30,10)               & (6,6)                               & 6                     \\
							
				EU       & (0,5)           & (10,10)               & (6,3)                               & 6                     \\
				
				\bottomrule
			\end{tabular}
			
		\label{tab:synthetic-datasets}
	\end{threeparttable}
\end{table}

The EEH dataset is a demonstration that, even in a dataset which have clusters of equal density and equal node degree, it can be difficult to separate the two clusters because there is a large number of connections between
the two clusters. 

The explanation is illustrated in Figure \ref{fig:EEH}. WL and WDK could not separate the two clusters as expected. But, mWDK is able to separate the two clusters at high $h$ because it could increase the similarity within each cluster while reducing the inter-cluster similarity.
\begin{figure}[ht]
	\centering
	\includegraphics[width=\linewidth]{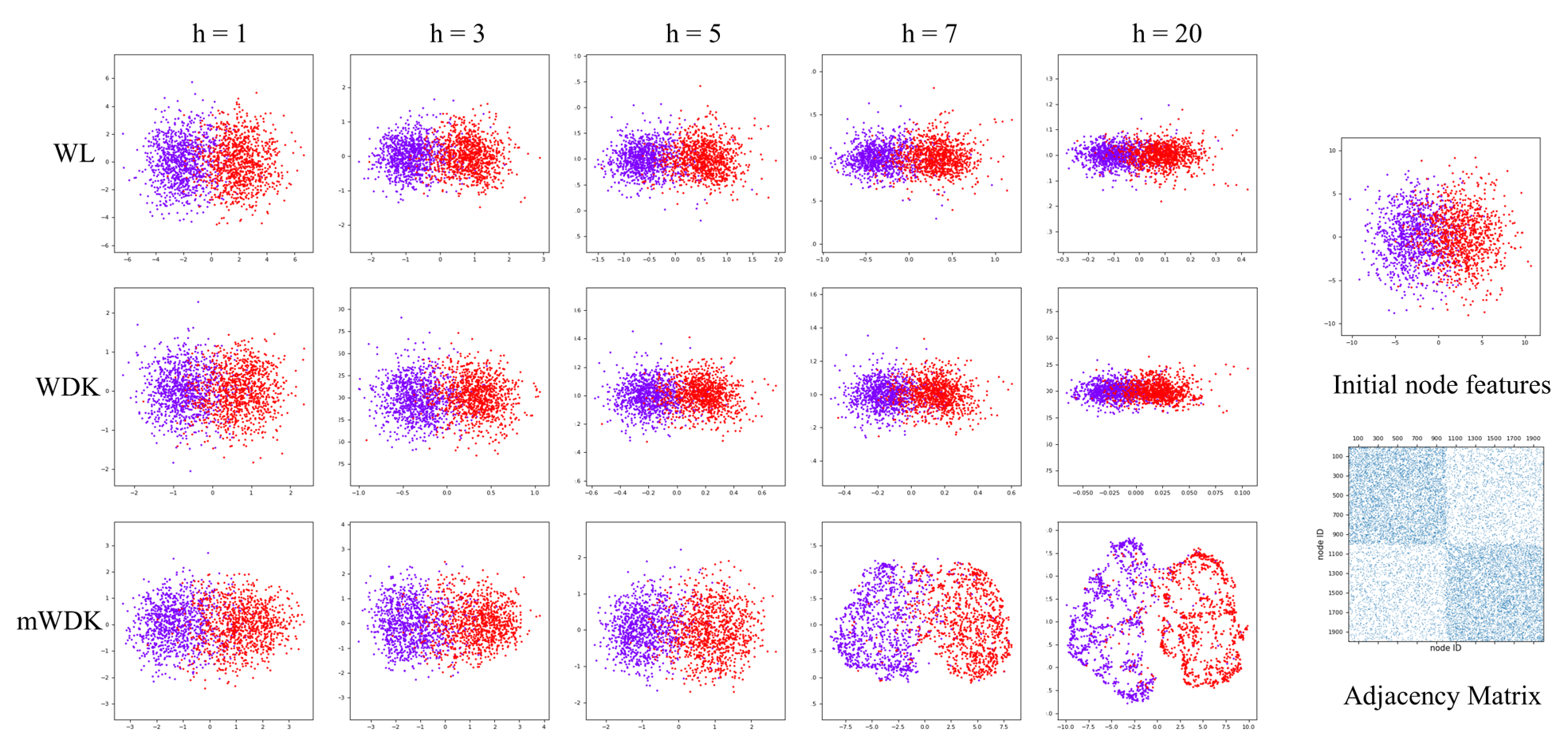}
	\caption{2D t-SNE Visualizations of the embedded spaces of WL, WDK, mWDK at various $h$ on  the EEH dataset.}
	\label{fig:EEH}
\end{figure}

\subsection{Proof of Theorem 1}

    The feature map of $\kappa_\psi$ with $t$ Voronoi tessellations \cite{DBSCAN-IK-AAAI2019} is defined as $\phi : x \rightarrow \{0,1\}^{t\times \psi}\  \forall x \in \mathbb{R}^d$,  where $\phi_{ij}(x) =1, \forall i \in [1,t]$ \& $\phi_{ik}(x) =0, \forall k \ne j$ for any $j \in [1,\psi]$; and $\parallel \phi(x) \parallel = \sqrt{t}$. Let $\ell_p$ be the L$_p$-norm. 
    
For any two points $x,y \in \mathbb{R}^d$, the total number of 1's in  $\phi(x)$ and $\phi(y)$ is $2t$; $\ell_{p}(\phi(x), \phi(y))^{p}$ denotes the number of unmatched 0 and 1 feature value-pairs of $\phi(x)$ \& $\phi(y)$; and $t\kappa_{\psi}(a, b)=<\phi(x),\phi(y)>$ denotes the number of matched 1-to-1 feature value-pairs.  Thus $\phi(x)$ \& $\phi(y)$ have a total number of 1's with matched 1-to-1 of $2t\kappa_{\psi}(x, y)$. As a result, we have:
\begin{equation}\label{eq:pair}
\ell_{p}(\phi(x), \phi(y))^{p}+2t \kappa_{\psi}(x, y) = 2t.
\end{equation}
If the two clusters satisfy Eq (10) in the paper, then $\kappa_\psi$ ensures that any $x \in \mathcal{C}_1$ and $y\in \mathcal{C}_2$ are not in the same Voronoi cell, resulting in $\kappa_{\psi}(x,y)=0$. 
Consequently, according to Eq (\ref{eq:pair}), maximum $\delta =\sqrt[p]{2t}$.

\subsection{Gaussian Kernel in mWDK}
\label{appendix: gk-mWDK}
\textbf{The use of Gaussian Kernel in mWDK cannot maintain well-separated clusters at each iteration}.
For Gaussian Kernel $\kappa_\sigma$ with bandwidth $\sigma$, it holds that 
\begin{equation}
    \begin{aligned}
    \ell_2(\phi(x),\phi(y)) &=\sqrt{\kappa_\sigma (x,x)+\kappa_\sigma (y,y)-2\kappa_\sigma (x,y)}\\
    &=\sqrt{(\frac{1}{\sqrt{2\pi }\sigma})^d[2-2e^{{-\ell_2(x,y)^2}/{2\sigma^2}}]}.
    \end{aligned}
\end{equation}
For two points $x,y$ in two different clusters, due to the data independence of Gaussian kernel, the kernel distance $\ell_2(\phi(x),\phi(y))$ is  dependent on $\ell_2(x,y)$ only. If $\ell_2(x,y)\geq \Delta$, we have\\$\ell_2(\phi(x),\phi(y))\geq \sqrt{(\frac{1}{\sqrt{2\pi}\sigma})^d[2-2e^{{-\Delta^2}/{2\sigma^2}}}]$.\\
Thus, for two points $x,y$ in two different clusters, the kernel distance of $\kappa_\sigma $ has a lower bound only, and there is nothing to prevent the lower bound being zero. When combined with WL, this lower bound can not guarantee cluster separation after the WL iteration.

 }

\end{document}